\documentclass[twocolumn,3p,authoryear]{elsarticle}
\usepackage[utf8]{inputenc}
\usepackage[T1]{fontenc}
\usepackage{stix} % change font 
\usepackage{amsmath} % equations
\usepackage{lineno} % line numbers
\usepackage{float} % float environment for figures
\usepackage[breaklinks=true]{hyperref}  % For clickable links
\usepackage[capitalize]{cleveref} % smart    references
\usepackage{xurl} % to break long urls
\usepackage{booktabs}
\usepackage[table]{xcolor}

\usepackage{caption}
\usepackage{dblfloatfix}

\hypersetup{colorlinks=true} % color links

\begin{document}

% color links
\hypersetup{
    citecolor=[rgb]{0.1, 0.3, 0.4},
    linkcolor=[rgb]{0.1, 0.3, 0.4},
    urlcolor=[rgb]{0.1, 0.3, 0.4},
}

\begin{frontmatter} 

\title{SERA-H: Super-Resolution of Sentinel Time Series for Fine-Scale Canopy Height Mapping}

% %% Authors
\author[Lsce]{Thomas Boudras}
\author[Lsce]{Martin Schwartz} %% Author name
\author[KU]{Rasmus Fensholt}
\author[KU]{Martin Brandt}
\author[Lsce,Kayrros]{Ibrahim Fayad}
\author[INRAE]{Jean-Pierre Wigneron}
\author[Lsce,Munster]{Gabriel Belouze}
\author[ENS]{Fajwel Fogel}
\author[Lsce]{Philippe Ciais} %% Author name

% %% Affiliations
\affiliation[Lsce]{
  organization={Laboratoire des Sciences du Climat et de l’Environnement (LSCE), CEA, CNRS, UVSQ,
Université Paris-Saclay},
  city={Gif-sur-Yvette},
  country={France}
}

\affiliation[Kayrros]{
    organization={Kayrros SAS},
    city={Paris},
    postcode={75009},
    country={France}
}

\affiliation[ENS]{
  organization={CNRS \& Département d'Informatique, École Normale Supérieure -- PSL},
  addressline={45 Rue d'Ulm},
  postcode={75005}, 
  city={Paris},
  country={France}
}

\affiliation[KU]{
    organization={Department of Geography and Geology, University of Copenhagen},
    addressline={Øster Voldgade 10},
    city={Copenhagen},
    postcode={DK-1350},
    country={Denmark}
}

\affiliation[INRAE]{
  organization={INRAE, Bordeaux Aquitaine Center},
  addressline={71 avenue E. Bourlaux, CS 20032},
  postcode={33882},
  city={Villenave d'Ornon},
  country={France}
}

\affiliation[Munster]{
  organization={Department of Information Systems, University of M\"unster},
  city={M\"unster},
  country={Germany}
}

% %% Abstract
\begin{abstract}

High-resolution mapping of canopy height is essential for forest management and biodiversity monitoring. Although recent studies have led to the advent of deep learning methods using satellite imagery to predict height maps, these approaches often face a trade-off between data accessibility and spatial resolution. To overcome these limitations, we present SERA-H, an end-to-end model combining a super-resolution module (EDSR) and temporal attention encoding (UTAE). Trained under the supervision of high-density LiDAR-derived Canopy Height Models (CHM), our model generates  2.5~m resolution height maps from freely available Sentinel-1 and Sentinel-2 (10~m) time series data. Evaluated against an open-source benchmark dataset in France, SERA-H, with a MAE of 2.6~m and R$^2$ of 0.82, not only outperforms standard Sentinel-1/2 baselines but also approaches the accuracy of methods based on commercial very high-resolution imagery (e.g., SPOT-6/7), while relying solely on freely available data. These results demonstrate that combining high-resolution ALS supervision with the spatio-temporal information embedded in Sentinel time series enables the reconstruction of spatial detail finer than the native resolution of the input imagery. By approaching the accuracy of costly commercial imagery, SERA-H opens the possibility of mapping temperate forests using publicly available data with high revisit frequency.

\end{abstract}

%% Keywords
\begin{keyword}

  Deep Learning \sep Forest Height \sep Time Series \sep Sentinel \sep ALS \sep Super-Resolution \sep Attention Encoder 

\end{keyword}

\end{frontmatter}

\section{Introduction}

Accurate canopy height mapping is essential for understanding forest ecosystems, estimating biomass and stored carbon, and monitoring forest dynamics \citep{lefsky_lidar_2002, goetz_lidar_2010, schwartz_forms_2023, tomppo_methods_2008, bossy_state_2025}. Such maps are valuable for sustainable forest management, biodiversity assessment, and climate change mitigation policies. 

Canopy height maps can be derived by processing data from Airborne Laser Scanning (ALS) and NASA's Global Ecosystem Dynamics Investigation (GEDI). Direct measurements are also obtained from National Forest Inventories (NFIs) at point locations. While the GEDI mission has established itself as the reference for monitoring the vertical structure of forests on a global scale, it operates by nature as a sampling mission providing sparse spatial and temporal samples (so-called footprints). ALS is the reference for high-resolution local canopy height mapping, but its deployment remains costly and geographically limited. As for NFI inventories, although they provide direct \textit{in-situ} measurements, they remain limited to a small number of plot locations and do not allow for continuous spatial monitoring. To overcome the discontinuity of these reference measurements, deep learning approaches have emerged as an effective solution. These approaches exploit the continuous spatial and temporal coverage of optical (e.g., Sentinel-2, Landsat, SPOT) and radar (e.g., Sentinel-1, PALSAR-2) imagery to predict canopy height maps under the supervision of LiDAR acquisitions, such as those from GEDI or ALS data. To achieve this, models leverage optical reflectance, which correlates with vertical forest structure \citep{woodcock_mapping_1994}, and radar backscatter, which depends on the geometric properties of the forest. However, these methods encounter saturation effects specific to each type of signal, which can limit their ability to make accurate predictions about large trees. Despite these constraints, this framework enables large-scale canopy height prediction using data that is both frequently updated and cost-effective.

Recently, several deep learning approaches have produced canopy height maps derived from open-access satellite data \citep{lang_high-resolution_2023, schwartz_retrieving_2025, pauls_capturing_2025}. However, the level of detail in these maps remains constrained by the spatial resolution of the source imagery (at best 10~m resolution with Sentinel-2). Therefore, these maps can at best go as far as the tree group level, but never into greater detail. To overcome these spatial limitations and to achieve the precision closest to the individual tree level, recent approaches have focused on using very high-resolution commercial imagery (e.g., Maxar \citep{tolan_very_2024}, PlanetScope \citep{liu_overlooked_2023}, SPOT-6/7 \citep{fogel_open-canopy_2025}). However, while these imagery sources offer superior spatial detail, they suffer from either restricted access, limited spatial coverage, or high costs.

Open access to ALS data has expanded significantly (LiDAR HD program \footnote{\url{https://geoservices.ign.fr/lidarhd}}, \citet{fischer_global_2025}) in recent years providing extensive coverage of high-quality data. Beyond serving as the most accurate reference for locally estimating canopy height \citep{alexander_influence_2018, lefsky_estimates_2005}, ALS datasets offer a very high spatial resolution (typically 1-2~m). On the other hand, the highest resolution provided by open-access satellite data comes from the Sentinel missions, part of ESA's Copernicus Earth Observation program. At 10~m, the resolution of Sentinel-1 and Sentinel-2 satellite images remains considerably coarser than the level of detail offered by ALS-based canopy height maps. To match the resolution of the input imagery with the reference data for model training, the most common approach is to downgrade the resolution of the canopy height derived from ALS data to align it with the satellite's native resolution \citep{liu_overlooked_2023, su_canopy_2025, su_fused_2025}, inevitably resulting in significant information loss. Here, we propose an alternative approach: to increase the resolution of Sentinel images to match the fidelity of ALS-derived canopy height maps.

To increase the spatial resolution of images, a solution is the application of super-resolution, a method for reconstructing high-resolution images from lower-resolution data. Super-resolution techniques are widely applied in urban remote sensing: Several building segmentation models employ a two-step sequential pipeline, in which a super-resolution deep learning model first upsamples the image, serving as a preprocessing step for a separate model dedicated to building detection \citep{chen_large-scale_2023, zhang_making_2021}. End-to-end architectures have also been implemented, integrating super-resolution and segmentation within a unified network in order to simultaneously optimize both tasks \citep{ayala_pushing_2022}.

At the same time, a recent study has demonstrated the value of directly exploiting the time series of raw Sentinel-2 images, in contrast to the standard practice of using annual or seasonal mosaics, to estimate canopy height \citep{pauls_capturing_2025}. By avoiding the loss of information inherent in annual compositing, this model exploits the seasonal and intra-annual dynamics of input Sentinel-2 images, significantly improving the accuracy of height predictions, particularly for tall trees. Another study by \citet{sirko_high-resolution_2023} combined a super-resolution approach and multi-temporal images, by simultaneously using spatial and temporal variations in successive acquisitions to produce higher-resolution results in various urban remote sensing contexts. This model has proven effective for tasks such as building segmentation, centroid and height estimation, and road segmentation. Fundamentally, this performance relies on exploiting the sub-pixel shifts inherent in revisited acquisitions, which allows the model to resolve aliasing and reconstruct fine spatial details by leveraging this temporal redundancy. 

Current methods lack the capacity to generate canopy height maps at a near-individual tree scale without relying on costly, restricted, or infrequent commercial imagery. To address this limitation, we introduce SERA-H (Super-resolution for Environmental Rapid Analysis - Height), an end-to-end deep learning framework designed to generate very high-resolution (2.5~m) canopy height maps directly from coarser Sentinel-1 and Sentinel-2 time series (10~m). 
Our approach aims to fully leverage the input data to reconstruct canopy structure at a finer resolution than the native Sentinel grid, thereby avoiding the information loss typically induced by downsampling ALS reference data. We set this target at 2.5~m, chosen to come as close as possible to the 1.5~m ALS reference. To achieve this, we combine a super-resolution module with a temporal attention mechanism designed to process satellite image time series. This article first details the proposed architecture (\cref{sec:method}). We then present an ablation study to validate our design choices (\cref{sec:ablation_results}), before benchmarking our results against recent state-of-the-art methods (\cref{sec:sota_comparison_results}).

\section{Materials and Methods}
\label{sec:method}

\subsection{Data}
\label{sec:data}

The proposed approach aims to transform Sentinel-1 and Sentinel-2 time series (10~m) into high-resolution canopy height maps (2.5~m) using super-resolution and temporal regression. This section details the datasets used for training and evaluation. We first describe the Open-Canopy benchmark (\cref{sec:dataset}), which provides the ALS-based reference height maps, the dataset splits, and the vegetation mask. Then, we present the acquisition and preprocessing of the input Sentinel satellite time series (\cref{sec:satellite_data}).

\subsection{Open-Canopy Dataset: Reference Data, Splits, and Vegetation Mask}
\label{sec:dataset}

To train and evaluate SERA-H, we used the Open-Canopy dataset \citep{fogel_open-canopy_2025}, a public benchmark specifically designed for high-resolution canopy height estimation. Derived from the French LiDAR HD campaigns acquired between 2021 and 2023\footnote{\url{https://geoservices.ign.fr/lidarhd}}, the Open-Canopy dataset provides ALS-based canopy height maps at 1.5~m resolution which served as reference for height prediction. For the purpose of our study, we downsampled these reference ALS-based canopy height images to 2.5~m using max-pooling aggregation. Max-pooling was chosen because it preserves the top-of-canopy height within each aggregated cell, retaining the dominant canopy structure that the model is trained to predict. 

This 2.5~m target resolution corresponds to a $\times 4 $ upsampling of the 10~m Sentinel grid. This choice reflects a trade-off between two opposing constraints. On the one hand, a $ \times 2$ factor (5~m) would not come close enough to the individual-tree scale, with each pixel still covering 25~m$^2$. Moreover, it would require additional aggregation of the 1.5~m reference, thereby discarding more of its information. This runs counter to the very purpose of our use of super-resolution, which is precisely to preserve that information. On the other hand, a $\times 8 $ factor (1.25~m) would fall below the resolution of the reference itself (1.5~m). More generally, the Sentinel input carries limited fine-scale information, constrained by its 10~m native resolution and by a geolocation accuracy of the order of several meters. The finer the target, the further it moves beyond the actual information content that the Sentinel input can provide, so that the reconstructed detail relies increasingly on the LiDAR-conditioned prior, increasing the risk of hallucinated or spurious detail. In this study, we therefore considered the $\times 4 $ factor to offer the best balance between fine detail and reliability.

\begin{figure}[htb]
    \centering
    \includegraphics[width=\linewidth]{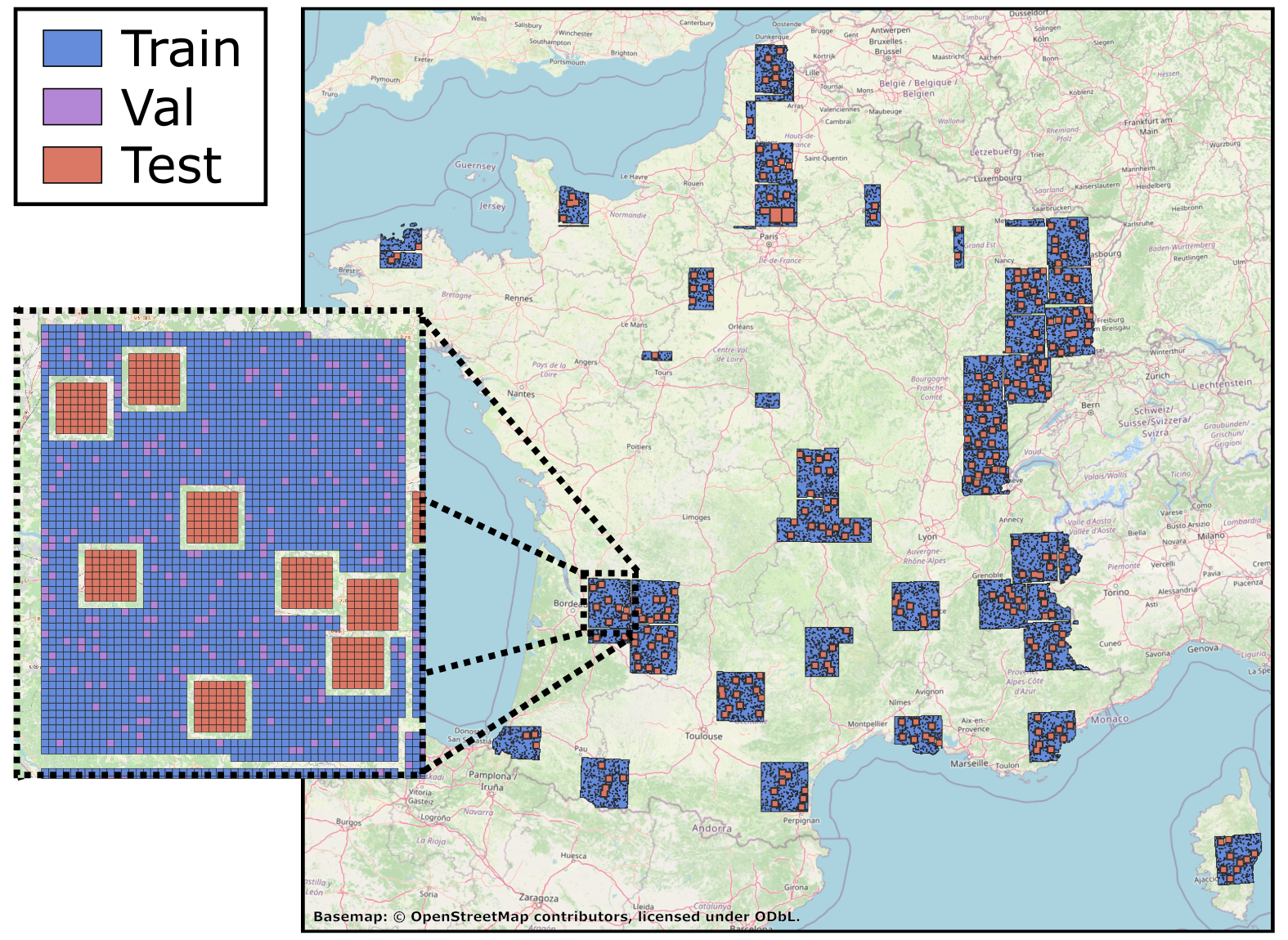}
    \caption{\textbf{Spatial distribution of the Open-Canopy dataset.} The map shows the distribution of the dataset across France \citep{fogel_open-canopy_2025}. The inset zoom on a region of this dataset shows the grid-based partition where each small grid square represents an individual 1 $\times$ 1~km tile. These tiles are divided between the training (blue), validation (purple), and test (red) tile sets. To prevent spatial data leakage, the test tiles are grouped together into larger blocks  and a 1~km exclusion buffer is systematically applied around these aggregated test areas.}
    \label{fig:figure_1_dataset}
\end{figure}

The dataset spans 87,383 tiles of $1\times1$~km distributed across France, ensuring a wide diversity of forest types and topographies. We adhered to the fixed spatial partition defined by Open-Canopy: 66,339~km$^2$ for training, 7,369~km$^2$ for validation, and 13,675~km$^2$ for testing (\cref{fig:figure_1_dataset}), with a 1~km buffer systematically applied around test areas to prevent spatial leakage. Validation tiles are used to monitor performance and adjust model hyperparameters during training, and test tiles are used to perform a final evaluation of model performance. 

Finally, to ensure that performance metrics focused on relevant areas, we restricted the evaluation to vegetation-covered zones using the vegetation mask provided by the Open-Canopy dataset. This mask was constructed by the union of a LiDAR-derived mask (height $>1.5$~m) and IGN forest polygons\footnote{\url{https://geoservices.ign.fr/bdforet}}. Note that this mask was exclusively applied during the validation and test phases for metric calculation and was not used during the training phase.

\subsubsection{Satellite Data: Sentinel-1 and Sentinel-2}
\label{sec:satellite_data}

Unlike the Open-Canopy benchmark which relied on commercial SPOT-6/7 imagery, our model deployed input imagery from the Sentinel-1 and Sentinel-2 sensor systems, key components of the European Space Agency's (ESA) Copernicus Earth observation program. These satellite systems provide complementary radar and optical data with global coverage and high temporal resolution. 

\begin{figure}[htb]
    \centering
    \includegraphics[width=\linewidth]{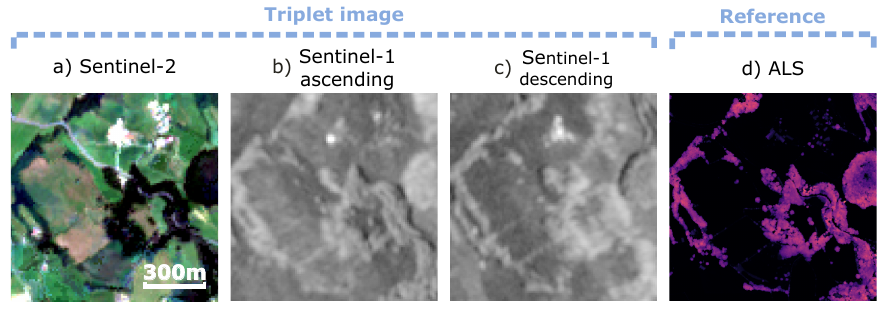}
        \caption{\textbf{Examples of input and reference data for a patch from the dataset.} (a) Sentinel-2 image, (b) Sentinel-1 ascending image, and (c) Sentinel-1 descending image form one triplet of the time series used as model input. (d) The corresponding reference is a canopy height map from the Open-Canopy dataset, resampled at 2.5~m. For visualization purposes, only a subset of the input bands is shown here (RGB for Sentinel-2 and VV for Sentinel-1). However, each triplet used by the model comprises 14 channels: 10 bands from $S2$, 2 from $S1_{\text{asc}}$ (VV/VH), and 2 from $S1_{\text{dsc}}$ (VV/VH).}
    \label{fig:figure_2_image_example}
\end{figure}

For Sentinel-2, we utilized the Harmonized Sentinel-2 Level-2A Surface Reflectance product\footnote{Available at: \url{https://developers.google.com/earth-engine/datasets/catalog/COPERNICUS_S2_SR_HARMONIZED}}. We selected 10 spectral bands specifically for their sensitivity to forest structural and biochemical properties: the visible RGB channels (B2, B3, B4) provide information on canopy texture and shadows, the vegetation red-edge (B5, B6, and B7) and near-infrared (NIR, B8 and B8A) bands help capture leaf area index and vegetation density, and the short-wave infrared (SWIR, B11 and B12) bands offer insights into canopy moisture and biomass. Regarding Sentinel-1, we used the C-band Synthetic Aperture Radar (SAR) Ground Range Detected (GRD) product\footnote{Available at: \url{https://developers.google.com/earth-engine/datasets/catalog/COPERNICUS_S1_GRD}} and we extracted both VV and VH polarizations. While VH polarization is primarily sensitive to volume scattering within the branches and foliage, VV polarization provides complementary information on vertical structures and surface interactions. We further integrated both ascending (south-to-north) and descending (north-to-south) orbits to provide the model with opposite viewing angles (\cref{fig:figure_2_image_example}b and \cref{fig:figure_2_image_example}c). While providing additional information, this dual-orbit perspective helps minimize geometric distortions and radar shadows. Altogether, since these diverse spectral and backscatter signals from Sentinel-2 and Sentinel-1 are fundamentally correlated with tree height, the network learns to interpret their complex non-linear relationships to transform these multi-modal features into a canopy height prediction.

We retrieved Sentinel images using the \texttt{geefetch} package\footnote{\url{https://geefetch.readthedocs.io/en/latest/}}  \citep{belouze_geefetch_2025}, which interfaces with the Google Earth Engine (GEE) API\footnote{\url{https://earthengine.google.com/}}. For each tile of the Open-Canopy Dataset, we collected all Sentinel images available within a 120-day window centered on the acquisition date of the corresponding ALS data. To minimize the influence of atmospheric perturbations, Sentinel-2 data were subjected to a cloud-masking procedure. Images with more than 40\% cloudy pixels or 20\% high-probability cloud coverage were excluded. Individual pixels with a cloud probability above 40\% or flagged in Sentinel-2's QA60 band were masked.  

Within this 120-day window, each Sentinel-2 ($S2$) image was paired with the temporally closest Sentinel-1 ascending ($S1_{\text{asc}}$) and descending ($S1_{\text{dsc}}$) acquisitions to form a triplet image $(S2, \allowbreak S1_{\text{asc}}, \allowbreak S1_{\text{dsc}})$ (\cref{fig:figure_2_image_example}). Therefore, each triplet image comprises 14 channels: 10 for Sentinel-2 bands, 2 for Sentinel-1 ascending (VV/VH), and 2 for Sentinel-1 descending (VV/VH). All these bands are harmonized at a resolution of 10~m using nearest-neighbor resampling. By forming all possible triplets from the available Sentinel images, we create a corresponding time series of triplet images for each tile. Across the entire dataset, the average length of each time series is 13.51 triplets per tile of the Open-Canopy dataset. Apart from the processing described above, no additional filters were applied to the Sentinel images. The goal is to enable the model to learn how to filter out noise, such as atmospheric artifacts or speckle, by leveraging the temporal redundancy of the time series.

\subsection{Model}
\label{sec:model}
\begin{figure*}[htb]
    \centering
    \includegraphics[width=\linewidth]{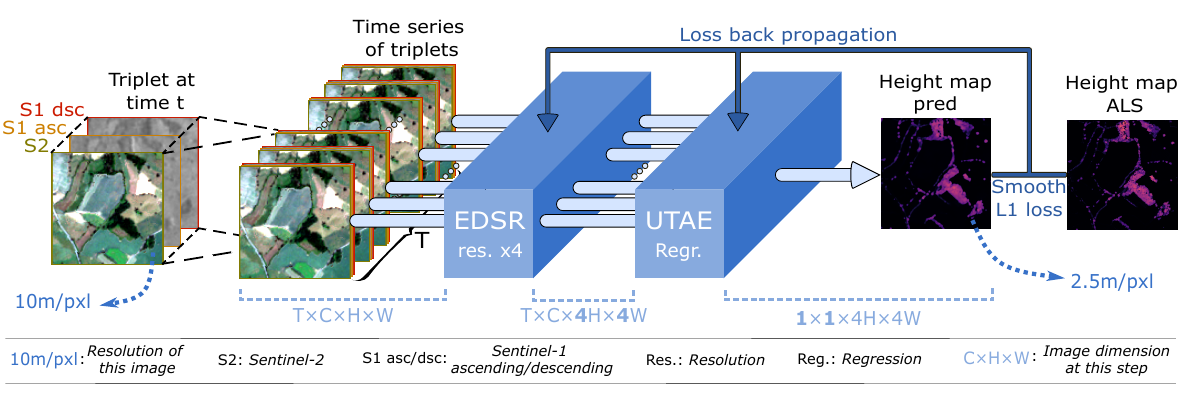}
    \caption{\textbf{Overview of the SERA-H model workflow.} Each input sample consists of a temporal sequence of triplets, where each triplet is composed of one Sentinel-2 optical image and its temporally closest Sentinel-1 ascending and descending acquisitions $(S2, S1_{\text{asc}}, S1_{\text{dsc}})$. This yields an input tensor of size $T \times C \times H \times W$, where $T$ is the fixed number of triplets in the temporal sequence, $C$ the total number of spectral and radar channels per triplet, and $H$ and $W$ the spatial dimensions of each image. The sequence is first processed by the EDSR super-resolution module, which upsamples all channels by $\times 4$ to match the reference ALS resolution. The super-resolved images are then passed through the UTAE encoder-decoder, which compresses the full temporal sequence using a temporal attention mechanism and outputs a single high-resolution canopy height map ($1 \times 1 \times 4H \times 4W$). Training is performed end-to-end using a Smooth L1 loss against ALS reference data, with gradients propagating through both UTAE and EDSR.}
    \label{fig:figure_3_model}
\end{figure*}

The SERA-H architecture is composed of two primary modules: a super-resolution module (EDSR, \cref{sec:edsr}) and a regression module (UTAE, \cref{sec:utae}). In the following subsections, we first detail the characteristics of each component individually, and then describe the global end-to-end workflow of the model (\cref{sec:overall_architecture}). We present the full workflow in \cref{fig:figure_3_model}.

\subsubsection{Enhanced Deep Super-resolution (EDSR) Module}
\label{sec:edsr}

Super-resolution aims to reconstruct a high-resolution image from low-resolution acquisitions by attempting to restore missing information. However, due to the very nature of this task, super-resolution algorithms are prone to creating hallucinations, i.e., non-existent details. These algorithms are systematically confronted with the perception-distortion trade-off \citep{blau_perception-distortion_2018}: the mathematical impossibility of simultaneously optimizing digital fidelity (proximity to actual pixel values) and visual realism (perceived sharpness). With a view to producing accurate canopy height maps, we have prioritized measurement accuracy and the reduction of these artifacts. We have therefore selected the EDSR (Enhanced Deep Super-Resolution) architecture, which has demonstrated excellent reconstruction performance in terms of fidelity to real values \citep{blau_perception-distortion_2018, karwowska_using_2022, ren_superbench_2025}. 

Derived from the ResNet architecture, EDSR optimizes the model by removing unnecessary normalization layers. This simplification allows for a much larger and more powerful network to be built. We thus adopted this architecture in our workflow to upsample each individual image within the time series, increasing the spatial resolution by a factor of 4.

In our approach, EDSR is used solely for its upsampling capability: it produces super-resolved feature maps that feed the subsequent regression module, without any image-reconstruction loss computed on these outputs. It therefore receives no image-space supervision and functions as a learnable upsampling module rather than a standalone super-resolution model.

\subsubsection{Regression Module: U-Net with Temporal Attention Encoder (UTAE)}
\label{sec:utae}

The second module is a regression network that ingests the super-resolved feature time series produced by the EDSR module and collapses it into a single canopy height map. For this purpose, we selected the UTAE (U-Net with Temporal Attention Encoder) \citep{garnot_panoptic_2022} for its temporal attention mechanism, which is well suited to fusing a complete image time series into a single prediction. This architecture combines a standard U-Net backbone \citep{ronneberger_u-net_2015} for spatial analysis with a dedicated temporal attention mechanism. Specifically, each image in the series is first encoded independently by the same convolutional encoder. Then, this temporal sequence is fused using a temporal attention mechanism that computed relevance masks to collapse the time dimension into a single representative embedding. This fused representation is finally decoded to reconstruct the high-resolution canopy height map.

\subsubsection{Overall Architecture}
\label{sec:overall_architecture}

Our approach relied on the synergy between super-resolution and multi-temporal regression to produce canopy height maps with a spatial resolution finer than that of the native Sentinel imagery. By exploiting the sub-pixel shifts inherent in revisited acquisitions, and guided by the high-resolution ALS reference, the model is trained to reconstruct finer forest structural details. To achieve this, we implemented an end-to-end architecture, enabling the simultaneous optimization of both the super-resolution (EDSR) and multi-temporal regression (UTAE) modules. By integrating these tasks into a single optimization step, we overcame the limitations of sequential two-stage pipelines, ensuring that the upsampling process was driven by the final canopy height prediction rather than by an intermediate reconstruction metric. The complete model contains 44.2 million trainable parameters, predominantly allocated to the EDSR module (97.5\%) compared to the UTAE module (2.5\%). The pipeline is illustrated in \cref{fig:figure_3_model} and described in detail below.

\paragraph{\textbf{Input Representation}}
The input data is formatted into a tensor $e \in \mathbb{R}^{B \times T \times C \times H \times W}$, where $B$ represents the batch size, $T$ the fixed sequence length of the time series, $C$ the number of channels, and $(H, W)$ the spatial dimensions. In our experiments, the shape was $2 \times 16 \times 14 \times 64 \times 64$.

\paragraph{\textbf{Super-Resolution Module (EDSR)}}
First, the images in the temporal sequence are independently upsampled. To do so, the batch and temporal dimensions are collapsed to form a flattened tensor $e^f \in \mathbb{R}^{(B \cdot T) \times C \times H \times W}$. This tensor is input into the EDSR model, which performs a $\times 4$ spatial upsampling, resulting in the super-resolved output $\hat{e}^f$:
\begin{equation}
    \hat{e}^f = \text{EDSR}(e^f) \in \mathbb{R}^{(B \cdot T) \times C \times H' \times W'}
\end{equation}
where $H' = 4H$ and $W' = 4W$. Finally, the output is reshaped to restore the original temporal structure, producing the tensor $\hat{e} \in \mathbb{R}^{B \times T \times C \times H' \times W'}$.

\paragraph{\textbf{Regression Module (UTAE)}}
Then, the super-resolved sequence is forwarded to the UTAE regression model. The spatial encoder processes each image, after which the temporal encoder applies an attention mechanism to collapse the temporal dimension. By utilizing the day of the year associated with each acquisition, represented via sinusoidal functions, the model computes an attention mask to weight pixels based on their date, reducing the sequence to a single representative feature map. Finally, the spatial decoder reconstructs the canopy height map $\hat{h}$:
\begin{equation}
    \hat{h} = \text{UTAE}(\hat{e}) \in \mathbb{R}^{B \times 1 \times H' \times W'}
\end{equation}
where each pixel value corresponds to the predicted canopy height.

\subsection{Implementation and Training Strategy}
\label{sec:implementation}

\subsubsection{Training Data Sampling}
During training, $1 \times 1$~km tiles were randomly sampled from the Open-Canopy dataset. The corresponding time series of $(S2, S1_{\text{asc}}, S1_{\text{dsc}})$ triplet images was retrieved. Tiles containing fewer than 4 distinct triplets were discarded. For the remaining tiles, as the model requires a fixed time series length of $T=16$, sequences shorter than this were padded by randomly duplicating existing triplets, while for longer sequences, we selected the $T=16$ triplets closest to the ALS acquisition date. The date assigned to each triplet corresponds to its Sentinel-2 acquisition date. Within the time series, triplets are then sorted by date. 
Finally, we cropped each triplet into $64 \times 64$ pixel patches to match the model's input dimensions. Corresponding ALS canopy height references were extracted at the same coordinates with a 2.5~m resolution, yielding $256 \times 256$ pixel reference images.

\subsubsection{Network Initialization and Optimization}
\label{sec:network_initialization_optimization}
The EDSR module was initialized using weights pre-trained on RGB images from the original study \citep{lim_enhanced_2017}. As the data used in our model had more channels (optical and radar), the input and output weights of the model were adapted by assigning the additional channels the average of the three RGB channel weights as initialization.

As it is an end-to-end architecture, both parts of the model (EDSR and UTAE) were trained together. The super-resolution module was therefore not frozen, but optimized simultaneously with the rest of the network, driven directly by the canopy-height loss. This aligns with its role as a learnable upsampling module, as introduced in \cref{sec:edsr}.

The batch size was set to 2, with each training sample consisting of a sequence of 16 triplet images $(S2, S1_{\text{asc}}, S1_{\text{dsc}})$, with each triplet image containing 14 channels. The Adam optimizer \citep{kingma_adam_2017} was used with a learning rate of $1 \times 10^{-4}$. The loss function employed was a Smooth L1 loss \footnote{\url{https://docs.pytorch.org/docs/stable/generated/torch.nn.SmoothL1Loss.html}}. A \texttt{ReduceLROnPlateau} \footnote{\url{https://docs.pytorch.org/docs/stable/generated/torch.optim.lr_scheduler.ReduceLROnPlateau.html}} scheduler was employed to reduce the learning rate by a factor of 0.1 whenever the validation loss failed to improve by at least 0.05 for 5 consecutive epochs. Early stopping was implemented to halt training if the validation loss failed to improve by at least 0.05 for 10 consecutive epochs. Training was performed on the Jean Zay supercomputer, using a node equipped with four NVIDIA Tesla V100 SXM2 32~GB GPUs and 40 Intel Cascade Lake 6248 CPUs.

\subsubsection{Inference Protocol}
During inference, for calculating metrics, predictions were made on patches of the same size as those used during training, i.e., $64 \times 64$ pixels. Inference was performed on the entire Open-Canopy test set by dividing these tiles into a grid of patches matching the model's input size. To limit edge effects, each $256 \times 256$ pixel inference patch was cropped by 21 pixels on each side, and an additional margin of 20 pixels was retained to create an overlap zone between adjacent patches. In these overlapping areas, predictions were averaged, which smoothed the transition between patches and produced a more homogeneous final map.

Inference was run on the same hardware as training (\cref{sec:network_initialization_optimization}), but using a single GPU and 10 CPU cores rather than the four GPUs and 40 CPU cores used for training. Predicting the entire 13,675~km$^2$ test set took approximately 9~h~40 min, i.e. about 2.5~s per km$^2$ at a batch size of 8. Regarding Sentinel data acquisition, our script and configuration achieved an average download rate of 3,580~km$^2$ per hour and an average storage of 6.4~MB per km$^2$. At this rate, mapping the whole of metropolitan France (about 552,000~km$^2$) would require roughly 153~h of data retrieval and 3.5~TB of storage, followed by about 390~h of inference on a single GPU. Since patches are processed independently, inference can be parallelized; hence using the four GPUs of the node would bring this down to about four days.

\subsection{Experimental Design}
\label{sec:experimental_design}

\subsubsection{Ablation Study Setup}
To evaluate the impact of different architectural components and identify the optimal configuration, we conducted an ablation study. Our proposed architecture, denoted as SERA-H (EDSR-UTAE (16img)), consists of the EDSR super-resolution module integrated with the UTAE regression model, utilizing a sequence of 16 input images. To reduce computational overhead during this comparative analysis, all models in the ablation study were trained on smaller input patches of $32 \times 32$ pixels.

To assess the contribution of the super-resolution module, we trained a variant replacing EDSR with a standard $\times 4$ bilinear interpolation upsampling, denoted as BI-UTAE$^*$ (16img). To ensure a fair comparison, the capacity of this variant was scaled to match the number of parameters of the full model by increasing the width of the UTAE encoder and decoder layers and expanding the attention mechanism dimensions. This guarantees that any performance gains are attributable to the architectural design rather than a simple increase in total capacity. This increase in the number of model parameters is indicated by the asterisk $^*$ appended to the model name.

To analyze the influence of temporal modeling, two additional models employing a U-Net architecture instead of UTAE were evaluated: BI-UNet$^*$ (bilinear interpolation upsampling) and EDSR-UNet (super-resolution upsampling). These models processed single composite images instead of full time series. These composites were generated by calculating the temporal mean for Sentinel-1 and the temporal median for Sentinel-2. Similarly, the number of parameters in the BI-UNet$^*$ model was increased to match that of EDSR-UNet to ensure a fair architectural evaluation.

To demonstrate the advantage of supervised learning at the target 2.5~m resolution instead of the native 10~m resolution, we trained a variant denoted as EDSR-UTAE-10m (16img). This model follows exactly the same architecture as SERA-H, except for the upsampling layer in the super-resolution module, which has been removed, yielding outputs at 10~m resolution. During training, the ALS reference resolution was downsampled to 10~m using max-pooling to match the model's output. For evaluation, the 10~m predictions were subsequently upsampled to 2.5~m using bilinear interpolation.

Finally, to determine sensitivity to sequence length, we trained two variants of SERA-H with reduced temporal length inputs: an eight-image model (EDSR-UTAE (08img)) and a four-image model (EDSR-UTAE (04img)).

\subsubsection{Comparison with State-of-the-Art Methods}
\label{sec:sota_comparison_presentation}

To evaluate the performance of SERA-H, we established a benchmarking protocol against five recent state-of-the-art (SOTA) methods: \citet{fogel_open-canopy_2025}, \citet{tolan_very_2024}, \citet{liu_overlooked_2023}, \citet{schwartz_retrieving_2025}, and \citet{pauls_capturing_2025}. In the remainder of this paper, these models are referred to as Fogel, Tolan, Liu, Schwartz, and Pauls, respectively. These baselines differ in their supervision strategies and input data resolutions. On one hand, Schwartz and Pauls trained models using GEDI data on Sentinel-1 and Sentinel-2 imagery, producing height maps at a 10~m resolution. While Schwartz used standard temporal composites, Pauls exploited Sentinel-2 time series, using composites only for Sentinel-1. On the other hand, approaches based on commercial very high-resolution imagery all used data derived from ALS as a supervision reference. For input data, Tolan used 1~m Maxar imagery, Liu utilized 3~m PlanetScope imagery, and Fogel employed 1.5~m SPOT-6/7 images. Regarding the latter, it is worth noting that while SPOT imagery is generally commercial, the specific dataset of satellite images used by Fogel is available free of charge for France through the Data Terra - DINAMIS \footnote{\url{https://openspot-dinamis.data-terra.org/}} institutional program. Furthermore, it is important to specify that Fogel was trained on the Open-Canopy benchmark as reference data, whereas the other ALS-supervised models relied on their own specific LiDAR datasets: Tolan focused on the USA, while Liu utilized a generalized dataset across Europe.

Regarding temporal alignment, these baseline maps display varying degrees of consistency with our 2021–2023 validation dataset. Fogel and Schwartz were fully aligned with the validation period (2021–2023), while Pauls covered the years 2021 and 2022. Conversely, the global maps based on commercial imagery relied on slightly older acquisitions: Liu provided a map for the fixed year 2019, and Tolan aggregated imagery primarily between 2018 and 2020. For these latter models, we used the product temporally closest to our reference dates, acknowledging that minor discrepancies due to forest dynamics (e.g., growth or harvest) might theoretically affect the comparison.
We quantify the magnitude of this temporal shift in \ref{app:temporal_shift}.

To ensure a consistent quantitative evaluation at our reference resolution of 2.5~m, a standardization procedure was applied to all baseline predictions. Maps with a native resolution lower than the target (Pauls, Schwartz 10~m and Liu 3~m) were upsampled using bilinear interpolation, whereas maps with a finer native resolution (Fogel 1.5~m and Tolan 1~m) were downsampled using max-pooling to preserve the top canopy height information.

\subsubsection{Evaluation Metrics}
\label{sec:evaluation_metrics}
The performance of the canopy height prediction model was evaluated using the Mean Absolute Error (MAE), Root Mean Squared Error (RMSE), Normalized Mean Absolute Error (nMAE), and Tree Cover IoU. All metrics, with the exception of Tree Cover IoU, were calculated only on pixels belonging to the vegetation mask described in \cref{sec:dataset}.

\paragraph{\textbf{Normalized Mean Absolute Error (nMAE)}}
To assess the relative accuracy of the predictions, we computed the Normalized Mean Absolute Error (nMAE), defined as:
\begin{equation}
    \text{nMAE} = \frac{1}{N} \sum_{i=1}^{N} \frac{|h_i - \hat{h}_i|}{h_i + 1}
\end{equation}
where $N$ is the total number of valid pixels, and $h_i$ and $\hat{h}_i$ denote the ALS reference and the predicted canopy height for pixel $i$, respectively. A regularization term of 1~m was added to the denominator to ensure numerical stability for low vegetation height. In addition to the vegetation mask, the nMAE was computed exclusively on pixels where the reference height exceeds 2~m, preventing the overestimation of errors in non-canopy areas.

\paragraph{\textbf{Tree Cover IoU}}
To evaluate the accuracy of forest canopy detection and the sharpness of predictions, we implemented the Tree Cover IoU metric. To do this, we generated two binary masks by applying a threshold of 2~m to the model predictions and reference data, above which we consider it to be the location of a tree. We then calculated the Intersection over Union (IoU) of these two ``tree cover'' masks, which gave us the final formula:
\begin{equation}
    \text{Tree Cover IoU} = \text{IoU}(M_{h}, M_{\hat{h}}) = \frac{|M_{h} \cap M_{\hat{h}}|}{|M_{h} \cup M_{\hat{h}}|}
\end{equation}
where $M_{h} = \{i \mid h_i \ge 2\}$ and $M_{\hat{h}} = \{i \mid \hat{h}_i \ge 2\}$ represent the binary vegetation masks for the reference and the model output, respectively. Although this indicator is only a derivative of the height threshold and does not provide an exact measurement of the canopy, it nevertheless allowed us to assess the extent to which the model correctly predicts the location of height areas.

\section{Results}
\label{sec:results}

\subsection{Ablation Study: Impact of Super-Resolution and Temporal Depth}
\label{sec:ablation_results}

Among all tested configurations, SERA-H (EDSR-UTAE with 16 images) achieved the best-performances, with the best scores across all metrics in the test dataset (\cref{tab:ablation_study}).

\begin{table*}[b]
    \centering
    \setlength{\tabcolsep}{10pt}
    \renewcommand{\arraystretch}{1.2}
    \caption{\textbf{Ablation study.} Evaluation of the impact of super-resolution (EDSR vs. Bilinear Interpolation, BI), temporal modeling (UTAE vs. U-Net), sequence length (4, 8, or 16 triplets), and supervision resolution (2.5~m vs. Sentinel native 10~m) on canopy height prediction. All metrics are evaluated in the test area at a standardized resolution of 2.5~m (model with a 10-meter output was upsampled for comparison). Note that to limit computational overhead, all models presented here were trained on reduced input patches of $32 \times 32$ pixels. \textit{Params} denotes the total number of trainable parameters in millions. Best results are highlighted in \textbf{bold}.}
    \begin{tabular}{l c c c c c}
        \toprule[1.5pt]
        \textbf{Model Configuration} & \textbf{Params} & \textbf{MAE} & \textbf{RMSE} & \textbf{nMAE} & \textbf{Tree Cover} \\
        & (M) & (m) & (m) & (\%) & \textbf{IoU} (\%) \\
        \midrule[1pt]
        BI-UNet$^*$ (mosaic) & 75.8 & 3.54 & 5.10 & 28.87 & 83.09 \\
        BI-UTAE$^*$ (16img) & 49.9 & 3.12 & 4.58 & 25.20 & 84.85\\

        \arrayrulecolor{black!30} 
        \midrule[0.1pt]  
        
        EDSR-UNet (mosaic) & 74.1 & 3.09 & 4.56 & 25.48 & 84.33 \\
        \midrule[0.1pt]  
        EDSR-UTAE-10m (16img) & 40.7 & 3.89 & 5.66 & 34.52 & 75.84 \\
        \midrule[0.1pt]  
        EDSR-UTAE (4img) & 44.2 & 3.04 & 4.49 & 24.89 & 84.68 \\
        EDSR-UTAE (8img) & 44.2 & 2.95 & 4.37 & 24.02 & 85.20 \\
        \midrule[0.1pt]  
        \arrayrulecolor{black}

        \begin{tabular}[c]{@{}l@{}} \textbf{SERA-H} \\[-0.5ex] \textbf{(EDSR-UTAE (16img))} \end{tabular} & 44.2 & \textbf{2.73} & \textbf{4.07} & \textbf{22.11} & \textbf{86.47} \\
        \bottomrule[1.5pt]
    \end{tabular}
    \label{tab:ablation_study}
\end{table*}

These results demonstrate that the super-resolution module substantially improves prediction quality compared to standard Bilinear Interpolation (BI). This trend is first observed in architectures based on composite mosaics: the EDSR-UNet model outperformed BI-UNet$^*$, reducing the Mean Absolute Error (MAE) by $0.45$~m (from $3.54$~m to $3.09$~m) and the Root Mean Square Error (RMSE) by $0.54$~m. This performance gain is confirmed for temporal architectures: replacing BI with EDSR within the UTAE model resulted in a significant MAE reduction of $0.39$~m (from $3.12$~m for BI-UTAE$^*$ to $2.73$~m for SERA-H). These findings confirm that a dedicated upsampling module enables the reconstruction of fine canopy details with significantly greater accuracy than conventional interpolation methods.

The use of the temporal dimension proved to be an equally important factor. Comparison of different architectures in \cref{tab:ablation_study} showed that models with time series (UTAE) consistently outperformed those based on static annual composites (U-Net). This efficiency was evident even in the configuration without super-resolution: the BI-UTAE$^*$ model performed better than BI-UNet$^*$ (MAE of $3.12$~m vs. $3.54$~m), while being notably lighter in terms of the number of parameters ($49.9$ million vs. $75.8$ million). The difference between these two methods is also visible on models that perform super-resolution: the SERA-H temporal model improved the MAE by $0.36$~m compared to its static equivalent EDSR-UNet ($2.73$~m vs. $3.09$~m). These results demonstrate that exploiting temporal redundancy does indeed allow for better characterization of forest structure than the use of static mosaics.

The significant deterioration in scores for the EDSR-UTAE-10m variant, which has the highest MAE in the benchmark at 3.89~m, demonstrates that degrading the resolution of ALS references leads to a major loss of structural details, confirming the value of high-resolution training targets. This performance gap also indicates that SERA-H performs true spatial reconstruction and is not simply a disguised version of a 10~m prediction followed by bilinear interpolation.

Finally, the ablation study revealed a direct positive correlation between the length of the input time series and the accuracy of the model. Reducing the number of images led to a gradual deterioration in performance. Specifically, the MAE increased by $0.22$~m when using 8 images (EDSR-UTAE (8img)) instead of 16 with SERA-H (from $2.73$~m to $2.95$~m), and worsened by an additional $0.09$~m when retaining only 4 images ($3.04$~m with EDSR-UTAE (4img)). This confirmed that the model benefited from dense temporal sampling to resolve ambiguities in canopy structure. Given that the average availability per area in our dataset was $13.51$ triplet images, the configuration of 16 triplets already exceeded this average. We therefore considered that exploring a model requiring an even longer sequence (e.g., 32 triplets) was not relevant.

Beyond these aggregate metrics, two complementary fine-scale analyses based on canopy edge and gap detection (\ref{app:fine_scale_metrics}) further support these conclusions, confirming that both the super-resolution module and the temporal series contribute to reconstructing structures finer than the Sentinel grid.

\subsection{Comparison with State-of-the-Art (SOTA)}
\label{sec:sota_comparison_results}

\subsubsection{Qualitative Evaluation}
\label{sec:qualitative_evaluation}

\begin{figure*}[t]
    \centering
    \includegraphics[width=\linewidth]{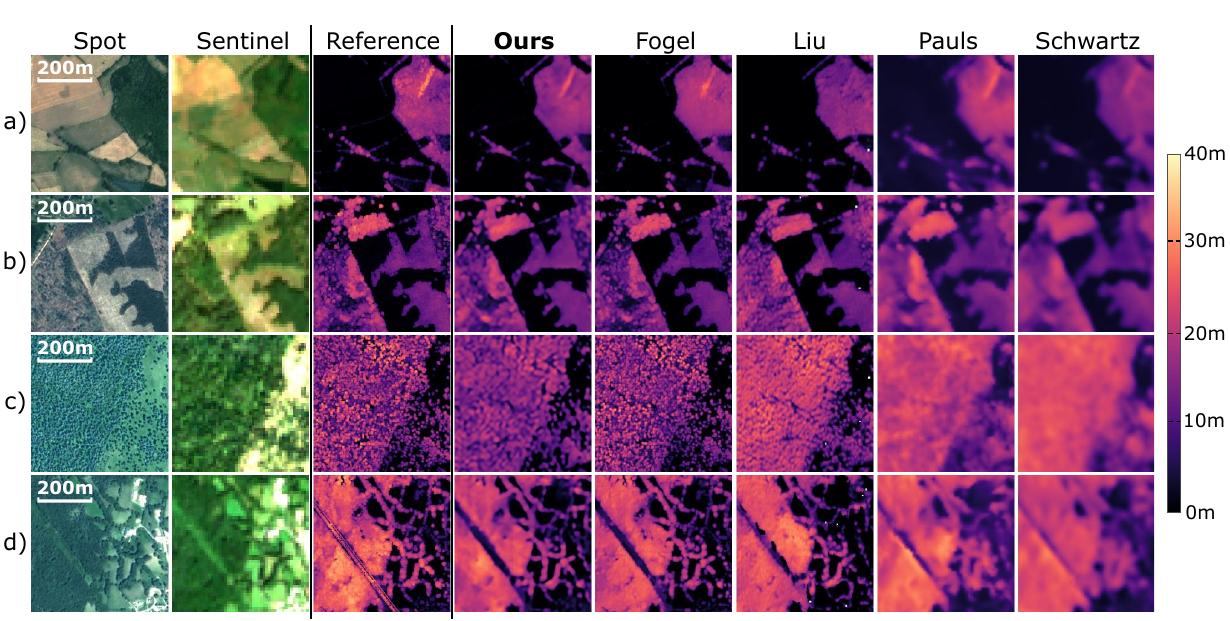}
    \caption{\textbf{Qualitative comparison of predicted canopy height maps.} Comparison of SERA-H with ALS reference data and four state-of-the-art canopy height maps (Fogel, Liu, Pauls, and Schwartz) over four selected study sites. The first two columns show the SPOT-6/7 and Sentinel-2 source images in RGB visualization. The third column displays the reference ALS-derived canopy height map at 2.5~m resolution. The following columns illustrate the canopy height estimates generated by each model. The study sites shown include: a) a chestnut grove in Aveyron ($44.202^\circ$N, $2.262^\circ$E), b) a maritime pine forest in Dordogne ($45.000^\circ$N, $0.309^\circ$E), c) a spruce forest in the Jura ($46.344^\circ$N, $5.991^\circ$E), and d) a beech forest in the Alps ($45.015^\circ$N, $5.793^\circ$E).}
    \label{fig:figure_4_qualitative_results}
\end{figure*}

Qualitative evaluation is essential to complement quantitative results and provide a visual understanding of canopy height predictions. \cref{fig:figure_4_qualitative_results} compares the predictions from SERA-H with the SOTA methods, showcasing the differences in predicted canopy structure across various forest types. Additional qualitative examples, randomly drawn from the test set, are provided in \ref{app:additional_examples}.

\paragraph{\textbf{Comparison with models trained on GEDI height}}
We acknowledge that a direct comparison of our model with those of Pauls and Schwartz may not be entirely equitable. These models were trained using sparse GEDI data as a reference, resulting in inherent discrepancies when evaluated against dense ALS measurements. Furthermore, their outputs are naturally less sharp due to the lower resolution of their target grids ($10$~m). The purpose of this comparison is therefore not to directly match these models, but rather to demonstrate typical achievements using Sentinel data and to highlight the potential of our approach using the same input imagery.

Despite utilizing the same coarse-resolution Sentinel data, SERA-H generated predictions that surpassed the native resolution of the input. It produced clearer and more precise contours of forests and individual trees, allowing us to approach the scale of the individual tree, which is significantly superior to other Sentinel-based products. This enhancement was largely driven by the use of ALS data as a reference, which provided a much higher frequency signal, enabling sharper and more accurate predictions. This confirms the utility of upsampling Sentinel images via a dedicated super-resolution method to align with the fine granularity of ALS data.

\paragraph{\textbf{Comparison with models trained on ALS products}}
Liu trained their model on a much broader dataset covering all of Europe; consequently, their model is not specifically tailored to the French forest types targeted in our test dataset. Despite this, Liu's model relies on commercial PlanetScope imagery, which has a resolution more than three times finer ($3$~m) than that of Sentinel. However, even against this higher-resolution input, SERA-H equals or even surpasses Liu's spatial definition (definition of forest and isolated trees). It can also be observed that, in many areas, SERA-H more accurately reflects internal variations in forest structure than Liu's predictions (variations in tree height and gaps in the forest).

The comparison with Fogel is the most relevant, as it is based on the same training dataset and ALS-based canopy height references (Open-Canopy dataset). Although Fogel employed SPOT-6/7 imagery, which offers a spatial resolution nearly seven times higher ($1.5$~m) than our Sentinel inputs, SERA-H achieved close prediction performance. While this resolution gap logically grants Fogel's maps a finer level of textural detail, the ability of SERA-H to approach these performance metrics demonstrates the effectiveness of our super-resolution approach in bridging the gap between open-access and commercial imagery.

\begin{table*}[b]
    \centering
    \setlength{\tabcolsep}{8pt}
    \renewcommand{\arraystretch}{1.2}
    \caption{\textbf{Quantitative comparison between SERA-H and state-of-the-art methods.} Our model is compared with Fogel, Liu, Tolan, Pauls and Schwartz, evaluated at a standardized resolution of 2.5~m. To match this resolution, finer maps were downsampled using max-pooling, while coarser maps were upsampled via bilinear interpolation. Input and reference images used for training each model are also specified in the columns 2 and 3, with their respective native resolutions provided in parentheses. Best results are highlighted in \textbf{bold}.}
    \begin{tabular}{l c c c c c c}
        \toprule[1.5pt]
        \textbf{Model} & \textbf{Input} & \textbf{Reference} & \textbf{MAE} & \textbf{RMSE} & \textbf{nMAE} & \textbf{Tree Cover} \\[-0.5ex] 
        & \textbf{Images} & \textbf{CHM} & (m) & (m) & (\%) & \textbf{IoU} (\%) \\
        \midrule[1pt]
        Pauls & S1-S2 (10m) & GEDI (10m) & 5.13 & 7.24 & 42.06 & 45.58 \\
        Schwartz & S1-S2 (10m) & GEDI (10m) & 4.47 & 6.17 & 45.96 & 35.12 \\

        \arrayrulecolor{black!30} 
        \midrule[0.1pt]           
        
        Tolan & Maxar (1m) & ALS (1m) & 5.49 & 7.52 & 41.74 & 70.15 \\
        Liu & Planet (3m) & ALS (3m) & 4.40 & 6.20 & 37.44 & 78.42 \\
        Fogel & SPOT-6/7 (1.5m) & ALS (1.5m) & \textbf{2.37} & \textbf{3.65} & \textbf{18.88} & \textbf{88.15} \\

        \midrule[0.1pt]           
        \arrayrulecolor{black}
        
        \textbf{SERA-H} & S1-S2 (10m) & ALS (2.5m) & 2.60 & 3.86 & 20.40 & 87.25 \\
        \bottomrule[1.5pt]
    \end{tabular}
    \label{tab:sota_comparison_results}
\end{table*}

\subsubsection{Quantitative Evaluation}
\label{sec:quantitative_evaluation}

\begin{figure*}[htb]
    \centering
    \includegraphics[width=0.9\linewidth]{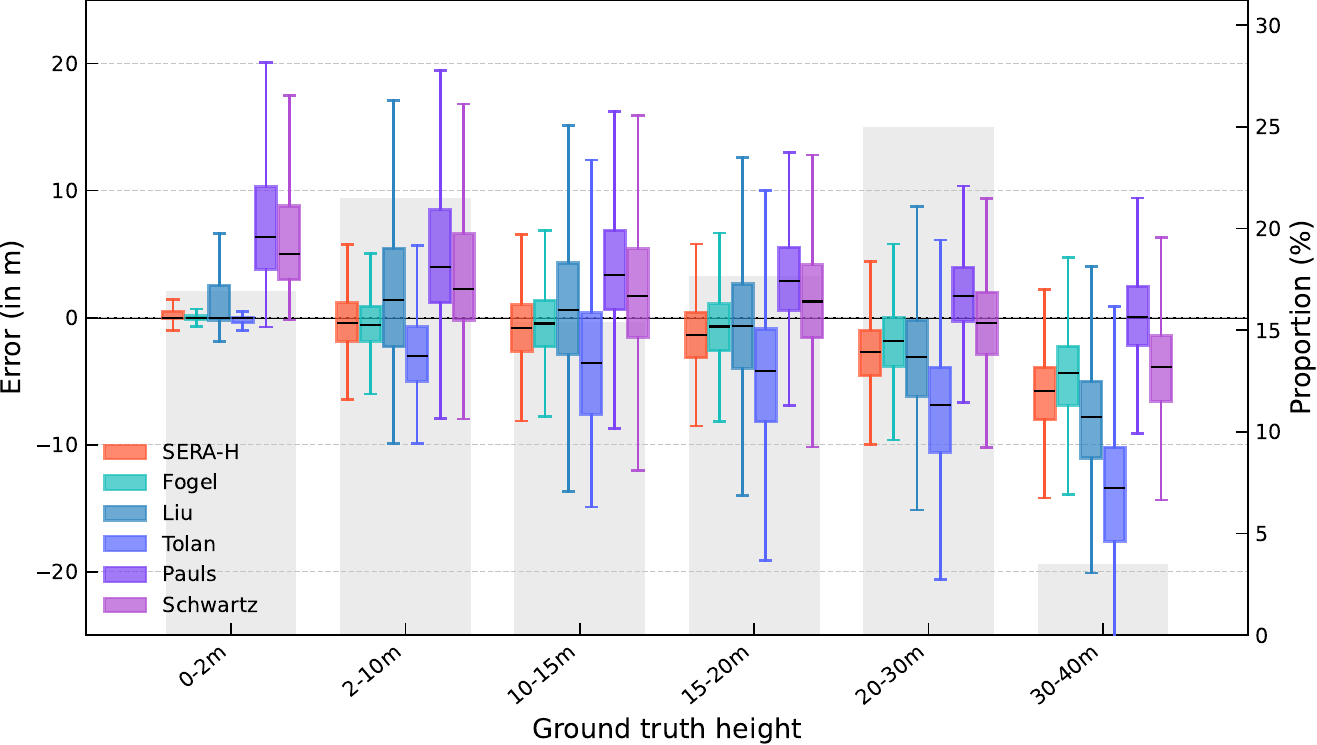}
    \caption{\textbf{Distribution of prediction error across reference canopy height bins.} The plot displays the error deviation ($\text{prediction} - \text{reference}$) for six models: SERA-H, Fogel, Liu, Tolan, Pauls, and Schwartz (primary y-axis, left). The shaded gray bars in the background represent the relative proportion of samples in each height class (secondary y-axis, right).}
    \label{fig:figure_5_boxplot}
\end{figure*}

\begin{figure*}[htb]
    \centering
    \includegraphics[width=0.9\linewidth]{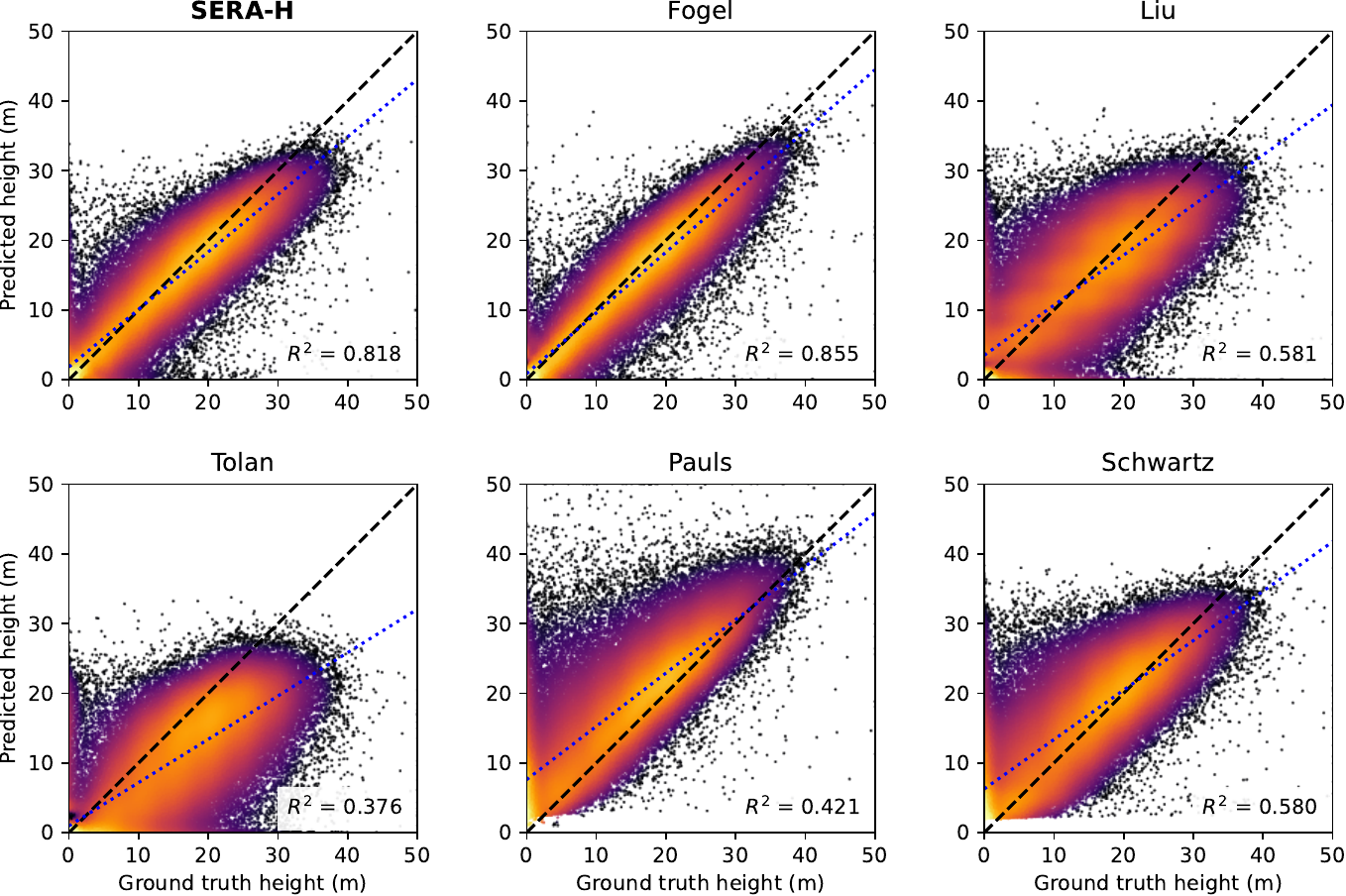}
    \caption{\textbf{Scatter density plots comparing predicted canopy heights with ALS reference data.} The panels display the performance of six models: SERA-H, Fogel, Liu, Tolan, Pauls, and Schwartz. The black dashed line indicates the perfect 1:1 fit, while the blue dotted line represents the linear regression. The coefficient of determination (R$^2$) is reported for each model. Note that for the Tolan model, a uniform noise in the range $[-0.5, 0.5]$ was applied to the original integer-valued predictions to enhance the visualization of point density; this adjustment does not affect the computed performance metrics. Brighter colors indicate a higher density of points.}
    \label{fig:figure_6_scatter_plot}
\end{figure*}

In this section, we present the quantitative evaluation of SERA-H alongside other state-of-the-art models. \cref{tab:sota_comparison_results} summarizes the performance metrics (detailed in \cref{sec:evaluation_metrics}) on the test dataset. To provide a more detailed analysis of the prediction errors, \cref{fig:figure_5_boxplot} illustrates the distribution of residuals across height bins, while \cref{fig:figure_6_scatter_plot} displays density scatterplots against the ALS reference data of the test dataset.

\paragraph{\textbf{Comparison with models trained on GEDI height}}
As noted in qualitative results (\cref{sec:qualitative_evaluation}), comparing our results with Pauls and Schwartz should be done with some caution, given their 10~m native resolution and reliance on GEDI reference. Metric discrepancies are therefore to be expected due to these differences in resolution and reference. However, as the height maps derived from ALS constitute the most reliable local estimate of canopy height at a very high resolution, it remains relevant to evaluate all models against this reference. Furthermore, this comparison allows us to identify the methodological benefit of SERA-H, revealing the accuracy improvement over existing methods that utilize Sentinel imagery as input.

The results reported in \cref{tab:sota_comparison_results} are consistent with the improved spatial detail observed in the qualitative results (\cref{sec:qualitative_evaluation}). SERA-H showed significantly lower errors than the Sentinel- and GEDI-based references: we obtained an MAE of $2.60$~m, compared to $4.47$~m for Schwartz and $5.13$~m for Pauls. The relative error (nMAE) was also reduced by half ($20.40\%$ compared to over $42\%$ for Pauls). Furthermore, with a Tree Cover IoU of $87.25\%$ compared to a range of $35$-$45\%$ for the other models, SERA-H demonstrated a much greater ability to correctly delineate vegetation. Finally, we verified these trends by calculating the metrics at a resolution of 10~m (\ref{app:10m_eval}), corresponding to the native resolution of Sentinel images and related products. Although performance differences were reduced with spatial aggregation, the hierarchy of results remained the same. For example, for MAE, SERA-H obtained $3.00$~m, while Pauls obtained $3.36$~m and Schwartz obtained $3.72$~m, confirming the robustness of our model's predictions.

Further intercomparison analysis (\cref{fig:figure_5_boxplot} and \cref{fig:figure_6_scatter_plot}) confirmed the good performance of the SERA-H model results. It highlights that models based on the Sentinel/GEDI pair systematically showed significantly greater error dispersion, as illustrated by more diffuse scatter plots and wider interquartile ranges. In addition, these sentinel/GEDI models show a strong tendency toward overestimation across most of the spectrum ($0$-$30$~m), where the median errors were above zero. However, their performance becomes broadly comparable to that of SERA-H within the 20 to 30~m class, and Pauls' model demonstrated a lower median bias for trees exceeding 30~m. While all models eventually encounter signal saturation as canopy height increases, the architecture proposed by Pauls appears to be the most resilient to this phenomenon. This greater resilience to saturation allows it to maintain more reliable predictions for tall trees, ultimately outperforming other models in this specific range. Nevertheless, SERA-H confirmed its overall superiority over models derived from Sentinel images across the rest of the height range.

\paragraph{\textbf{Comparison with models trained on ALS products}}
Quantitative analysis (\cref{tab:sota_comparison_results}) shows that SERA-H significantly outperforms the approaches of Tolan and Liu. Despite the theoretical advantage conferred by the use of very high-resolution commercial images (PlanetScope at 3~m resolution and Maxar at 1~m), these models yielded Mean Absolute Errors (MAE) of $4.40$~m and $5.49$~m, respectively, which were higher than the $2.60$~m obtained by SERA-H. This hierarchy is visually confirmed by the scatter plots (\cref{fig:figure_6_scatter_plot}): where the predictions of Liu and Tolan showed high dispersion and struggled to follow the 1:1 line (with R$^2$ values of $0.581$ and $0.376$), SERA-H showed strong linearity ($R^2 = 0.818$). To ensure that this gap does not merely reflect the temporal misalignment of the Liu and Tolan products (\cref{sec:sota_comparison_presentation}), we evaluated SERA-H under the same 2019 conditions (\ref{app:temporal_shift}). Although this shift degrades our metrics (MAE rising from 2.60 to 3.92~m), SERA-H still outperforms both models on all metrics, confirming that its advantage is not merely an artefact of date alignment. Beyond this temporal effect, the remaining discrepancy could be explained partly by domain shift, given that Tolan was trained in the United States and Liu across Europe, suggesting that excessive geographical generalization can be detrimental to local accuracy \citep{fischer_global_2025}.

The comparison with Fogel is particularly important, as it used exactly the same dataset and reference images (Open-Canopy) as compared to SERA-H. The fundamental difference lies in the input images: Fogel uses SPOT-6/7 images (1.5~m) while SERA-H uses Sentinel-2 (10~m), which is nearly seven times lower in resolution. Despite this resolution disadvantage, SERA-H managed to approach Fogel's performance (MAE of $2.60$~m vs. $2.37$~m). This good performance of SERA-H could also be seen in the graphs: the scatter plots were almost identical and the boxplots (\cref{fig:figure_5_boxplot}) revealed a similar error distribution, with medians close to zero and narrow interquartile ranges.

\section{Discussion}
\label{sec:discussion}

Our results indicate that a very high input resolution is not, by itself, sufficient to obtain accurate canopy height maps. Although Liu and Tolan rely on higher-resolution input imagery (3~m and 1~m, respectively), they do not outperform SERA-H on our French benchmark. This lower accuracy is at least partly attributable to domain shift, as these models were trained on different regions and reference datasets (the USA for Tolan, and a generalized European dataset for Liu). This suggests that a finer input resolution does not necessarily translate into better local accuracy unless the model is also trained on the local domain. Conversely, training on high-fidelity ALS reference data acquired over the target domain, while leveraging the abundant information provided by input time series, allows the model to reconstruct plausible fine structures at a resolution finer than the native resolution of the input imagery. 

These results suggest that SERA-H can be a competitive alternative to approaches based on commercial higher-resolution imagery. Although Fogel obtained slightly sharper boundaries through the use of higher-resolution satellite images, SERA-H comes close to this accuracy using only Sentinel-1 and Sentinel-2 data, which is publicly available worldwide. Furthermore, while SPOT images are generally available on an annual basis, the Sentinel-1 and Sentinel-2 satellites offer a significantly higher revisit frequency. This higher temporal density paves the way for more frequent updates, particularly for detecting abrupt, major disturbances occurring within a single year.

\begin{figure}[h]
    \centering
    \includegraphics[width=\linewidth]{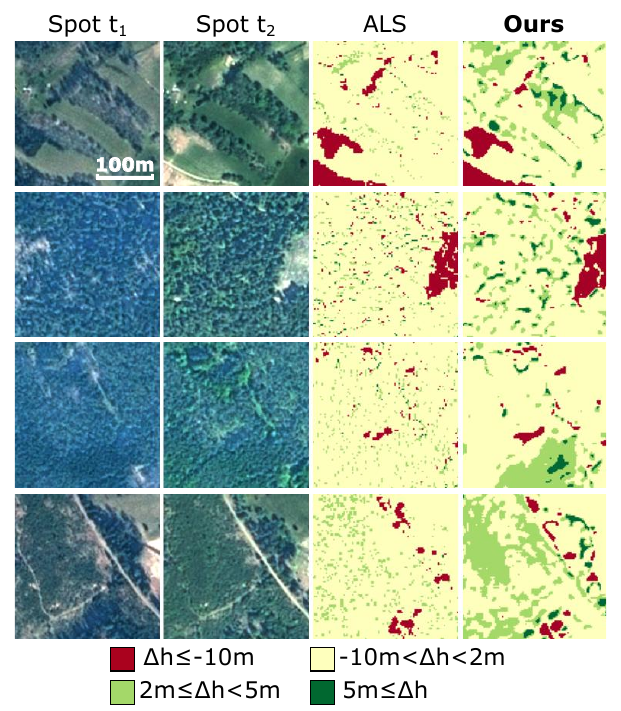}
    \caption{\textbf{Qualitative analysis of forest dynamics in the Vosges region (Winter 2020--2022).} The first two columns show SPOT-6/7 pansharpened images at 1.5~m resolution acquired on two separate dates ($t_1 = 2020$ and $t_2=2022$). The following columns illustrate the canopy height difference maps ($\Delta h = h_{t2} - h_{t1}$) derived respectively from the ALS reference and the SERA-H model predictions. Height variations are classified by thresholds: major losses (deforestation, $\le -10$~m) appear in red, stable or low variation areas ($-10$ to $2$~m) in yellow, moderate growth ($2$ to $5$~m) in light green, and strong growth ($> 5$~m) in dark green. All images are presented at the same scale.}
    \label{fig:figure_7_diff_height}
\end{figure}

\cref{fig:figure_7_diff_height} illustrates this potential with a few qualitative examples comparing height difference maps ($\Delta h = h_{t2} - h_{t1}$) obtained by ALS and by our model for forests in the Vosges region, France. For this comparison, the reference data was obtained from an initial IGN ALS campaign carried out in early 2020 \citep{ramirez_parra_how_2024} and the 2022 LiDAR HD campaign \footnote{\url{https://geoservices.ign.fr/lidarhd}}. By applying thresholds to identify major disturbance events, we observe that SERA-H seems to be able to identify and locate the main areas of deforestation (defined by a loss of at least $-10$~m). SERA-H could potentially serve as a monitoring or alert tool for clear-cutting, windfall, and other major losses, using fully accessible data.

However, analysis of these examples reveals a significant limitation concerning the detection of forest growth. The model struggles to accurately reproduce areas of moderate growth ($2$ to $5$~m) or strong growth ($> 5$~m) visible in the reference data. This result is consistent with the performance metrics: the natural growth of a temperate forest over a short period is often less than the model's Mean Absolute Error (MAE of $2.60$~m). As a result, the growth signal tends to be concealed by prediction noise (model uncertainty). Thus, while SERA-H shows real potential for detecting major forest disturbances, its use for quantifying changes of a few meters, such as gradual growth or small disturbances, will require further development, particularly with regard to the temporal stabilization of predictions.

\section{Conclusion and Limitations}
\label{sec:conclusion}

In this paper, we introduced SERA-H, a new end-to-end model combining super-resolution and spatio-temporal learning to generate very high-resolution (2.5~m) canopy height maps from coarser (10~m) Sentinel-1 and Sentinel-2 imagery. Our methodology was validated through a comprehensive ablation study, which confirmed the critical role of the super-resolution module (EDSR) and the temporal attention mechanism (UTAE) in reconstructing fine canopy details that standard interpolation methods fail to capture. Furthermore, we benchmarked SERA-H against five recent state-of-the-art methods, ranging from Sentinel-based models trained on GEDI \citep{schwartz_retrieving_2025, pauls_capturing_2025} to approaches leveraging commercial very high-resolution imagery supervised by ALS data \citep{fogel_open-canopy_2025, tolan_very_2024, liu_overlooked_2023}. The results demonstrate that, by leveraging the temporal density of open-access data, SERA-H outperforms other Sentinel-based models and surpasses commercial-imagery models penalized by domain shift (Liu, Tolan). It also approaches the accuracy of Fogel on the same benchmark, even though Fogel relies on commercial SPOT-6/7 imagery (1.5~m), with a spatial resolution nearly seven times higher than the freely available 10~m Sentinel data used here. This suggests that prediction accuracy can be, at least partially, decoupled from input resolution, positioning SERA-H as a potentially inexpensive tool for monitoring major forest disturbances in temperate forest.

Despite these significant advances, our method is subject to intrinsic physical and methodological limitations. While SERA-H significantly refines the predicted canopy structure, it remains constrained by the spectral information available in 10~m input pixels; consequently, it cannot fully reconstruct small isolated trees or fine details that are spectrally indistinguishable at this resolution, whereas native very high-resolution imagery might still capture these features. Furthermore, SERA-H remains sensitive to signal saturation, a physical constraint common to satellite-based height estimation. Similar to other state-of-the-art models, its accuracy declines for tall trees, particularly those exceeding 30~m, where the sensitivity of optical and radar signals to canopy height plateaus. A stratified analysis by height class and forest type (\ref{app:saturation_by_forest_type}) shows that this decline represents a systematic underestimation ($-5.9$~m median bias in the 30--40~m class) rather than an increase in random error, and that it affects all three forest types alike, confirming that this limit appears to be related to the Sentinel signal rather than to the forest type. Additionally, the model's performance is currently linked to the availability of high-density ALS data for supervision. The results obtained in this study were based solely on temperate forests in metropolitan France. Although the Open-Canopy reference dataset covers a wide variety of French forest types and topographies, the applicability of SERA-H to structurally different biomes remains to be demonstrated and transferring this approach to a new region would require high-density local reference data as well as a specific training phase. Future work should therefore focus on evaluating the model's generalization capabilities in data-scarce regions, such as tropical or boreal biomes. To mitigate the dependence on dense LiDAR campaigns, exploring transfer learning or weakly supervised techniques constitutes a promising direction for extending this approach to other areas.

\begin{table*}[b!]
    \centering
    \setlength{\tabcolsep}{8pt} 
    \renewcommand{\arraystretch}{0.9}
    \caption{Quantitative comparison between SERA-H (our model) and state-of-the-art methods (Fogel, Liu, Tolan, Pauls and Schwartz) evaluated at a standardized resolution of 10~m. All canopy height maps with finer native resolution were downsampled to 10~m using maximum pooling before metric calculation. Input and reference data sources are specified for each model in the column 2 and 3, with their respective native resolutions provided in parentheses.}
    \begin{tabular}{l c c c c c c}
        \toprule[1.5pt]
        \textbf{Model} & \textbf{Input} & \textbf{Reference} & \textbf{MAE} & \textbf{RMSE} & \textbf{nMAE} & \textbf{Tree cover} \\[-0.5ex] 
        & \textbf{images} & \textbf{CHM} & (m) & (m) & (\%) & \textbf{IoU} (\%) \\
        \midrule[1pt]
        Pauls & S1-S2 (10m) & GEDI (10m) & 3.72 & 4.90 & 23.19 & 55.71 \\
        Schwartz & S1-S2 (10m) & GEDI (10m) & 3.36 & 4.95 & 22.82 & 55.12 \\

        \arrayrulecolor{black!30} 
        \midrule[0.1pt]            
        
        Tolan & Maxar (1m) & ALS (1m) & 6.31 & 8.24 & 39.38 & 71.37 \\
        Liu & Planet (3m) & ALS (3m) & 4.09 & 5.66 & 28.79 & 81.43 \\
        Fogel & SPOT-6/7 (1.5m) & ALS (1.5m) & \textbf{2.44} & \textbf{3.62} & \textbf{15.81} & \textbf{89.95} \\

        \midrule[0.1pt]            
        \arrayrulecolor{black}
        
        \textbf{SERA-H} & S1-S2 (10m) & ALS (2.5m) & 3.00 & 4.11 & 18.65 & 89.35 \\
        \bottomrule[1.5pt]
    \end{tabular}
    \label{tab:sota_comparison_results_10m}
\end{table*}

\section*{Funding and Acknowledgements}

We are grateful to Jean-Pierre Renaud from the Office National des Forêts (ONF) for providing the 2020 Vosges ALS dataset. This work was supported by the One Forest Vision Initiative (OFVi), the AI4Forest project (ANR-22-FAI1-0002) and the PEPR FORESTT PC Monitor project (ANR-24-PEFO-0003). We also acknowledge support from the Danish National Research Foundation through the Center for Remote Sensing and Deep Learning of Global Tree Resources (TreeSense, DNRF192). This work was performed using HPC resources from GENCI-IDRIS (Grants 2025-AD010114718R1 and 2026-AD010114718R2).

\section*{Data Availability}

%The full implementation of SERA-H, including the code for Sentinel data retrieval, model training, and inference, is publicly available at \url{https://github.com/ThomasBoudras/SERA-H.git}

The Open-Canopy dataset used as reference in this study is available on Hugging Face at \url{https://huggingface.co/datasets/AI4Forest/Open-Canopy}.

\section*{Declaration of Generative AI in the Writing Process}

During the preparation of this work, the authors used Claude, Gemini and ChatGPT to assist in writing code and improving the manuscript's language and readability. After using this tool, the authors reviewed and edited the content as needed and take full responsibility for the content of the published article.

\appendix
\renewcommand{\thefigure}{\arabic{figure}}
\renewcommand{\thetable}{\arabic{table}}

\section{Evaluation at Standardized 10m Resolution}
\label{app:10m_eval}

While models such as Liu, Tolan, and Fogel benefit from high-resolution input imagery, Sentinel-based approaches (e.g., Pauls and Schwartz) natively produce canopy height maps at a coarser resolution (10~m) than the 2.5~m target. Evaluating these models at this finer scale requires upsampling them using bilinear interpolation, which inherently introduces artifacts and may unfairly disadvantage them. To ensure a fair comparison between models operating at different native resolutions, we conducted an additional evaluation standardized at 10~m. For this purpose, we resampled all maps with a finer resolution to a 10~m grid using a maximum pooling operation.

\begin{table*}[t]
    \centering
    \setlength{\tabcolsep}{10pt}
    \renewcommand{\arraystretch}{0.9}
    \caption{\textbf{Impact of temporal shift.} SERA-H predicted for 2019 is compared with the Liu and Tolan products, which likewise correspond to periods preceding the reference period, and with SERA-H predicted over the aligned 2021--2023 period. All models are evaluated at the standardized resolution of 2.5~m against the same 2021--2023 ALS references. Best results are highlighted in \textbf{bold}.}
\begin{tabular}{l c c c c c c}
        \toprule[1.5pt]
        \textbf{Model} & \textbf{Prediction} & \textbf{MAE} & \textbf{RMSE} & \textbf{nMAE} & \textbf{Tree Cover} & \textbf{ME} \\[-0.5ex]
        & \textbf{period} & (m) & (m) & (\%) & \textbf{IoU} (\%) & (m) \\
        \midrule[1pt]
        Tolan & 2018--2020 & 5.49 & 7.52 & 41.74 & 70.15 & -4.13 \\
        Liu & 2019 & 4.40 & 6.20 & 37.44 & 78.42 & -0.37 \\
        \textbf{SERA-H} & 2019 & 3.92 & 5.84 & 31.01 & 79.58 & -1.84 \\

        \arrayrulecolor{black!30}
        \midrule[0.1pt]
        \arrayrulecolor{black}

        \textbf{SERA-H} & 2021--2023 & \textbf{2.60} & \textbf{3.86} & \textbf{20.40} & \textbf{87.25} & -1.20 \\
        \bottomrule[1.5pt]
    \end{tabular}
    \label{tab:temporal_shift}
\end{table*}

The quantitative results presented in \cref{tab:sota_comparison_results_10m} confirm that Sentinel-based methods, such as Pauls and Schwartz, exhibit improved performance when evaluated at their native resolution of 10~m compared to finer scales. By avoiding upsampling artifacts, their Mean Absolute Errors drop to $3.72$~m and $3.36$~m, respectively.

However, despite this more favorable evaluation setting for standard Sentinel baselines, SERA-H maintains a significant performance lead across the dataset, achieving an MAE of $3.00$~m. This superiority is particularly evident in the spatial accuracy of the predictions, as reflected by the Tree Cover Intersection over Union (IoU). While Pauls and Schwartz achieve IoU scores of approximately $55$\%, suggesting difficulties in precisely delineating canopy boundaries, SERA-H reaches a Tree Cover IoU of $89.35$\%. This result highlights our model's ability to correctly localize vegetation, rivalling the performance of Fogel's high-resolution method ($89.95$\%) and demonstrating that SERA-H successfully reconstructs fine spatial details even from coarser input data.

\section{Impact of the Temporal Shift}
\label{app:temporal_shift}

As indicated in \cref{sec:sota_comparison_presentation}, the maps derived from commercial imagery are not perfectly aligned with our reference period (2021--2023): Liu provides a map for the year 2019, and Tolan aggregates acquisitions primarily between 2018 and 2020. Comparing these predictions with 2021--2023 references introduces a temporal shift, as forest growth or disturbances between the two dates may artificially widen the gap with the references. To assess the magnitude of this bias, we placed our own model under the same conditions: we predicted canopy height over the test-set areas with SERA-H, but this time from Sentinel images acquired within a four-month window centered on 1 July 2019. Then, we evaluated these predictions against the same 2021--2023 references. In addition to the metrics used throughout the paper, we also report the mean signed error (ME), i.e. the average signed difference between prediction and reference.

The temporal shift markedly degrades SERA-H's performance: its MAE rises from 2.60~m (2021--2023) to 3.92~m (2019), an increase of about 1.3~m attributable to the date misalignment alone. The temporal nature of this bias is confirmed by the mean signed error (ME): SERA-H already shows a slight underestimation under aligned conditions ($-1.20$~m, related to signal saturation on tall trees), which increases to $-1.84$~m for the 2019 prediction. This additional underestimation of about 0.6~m is consistent with forest growth between 2019 and the reference period.

It can also be noted that, even under this shift, SERA-H (2019) remains superior to Liu (2019) and Tolan (2018--2020) across all metrics: its MAE (3.92~m) stays below those of Liu (4.40~m) and Tolan (5.49~m), and it retains the best nMAE and Tree Cover IoU of the group. The ME values of these two models further reveal distinct behaviors: Tolan suffers from a strong systematic underestimation ($-4.13$~m), far larger than a temporal shift alone could explain, whereas Liu exhibits a near-zero bias ($-0.37$~m), its error stemming rather from dispersion. The temporal shift therefore accounts for only a limited part of these models' gap with SERA-H.

This experiment confirms that SERA-H's superiority over these commercial-imagery approaches cannot be reduced to an artefact of date alignment. However, this residual gap can be explained once again, at least in part, by the domain shift affecting these two models, as mentioned in the Discussion (\cref{sec:discussion}).

\begin{figure}[!b]
    \centering
    \includegraphics[width=0.5\textwidth]{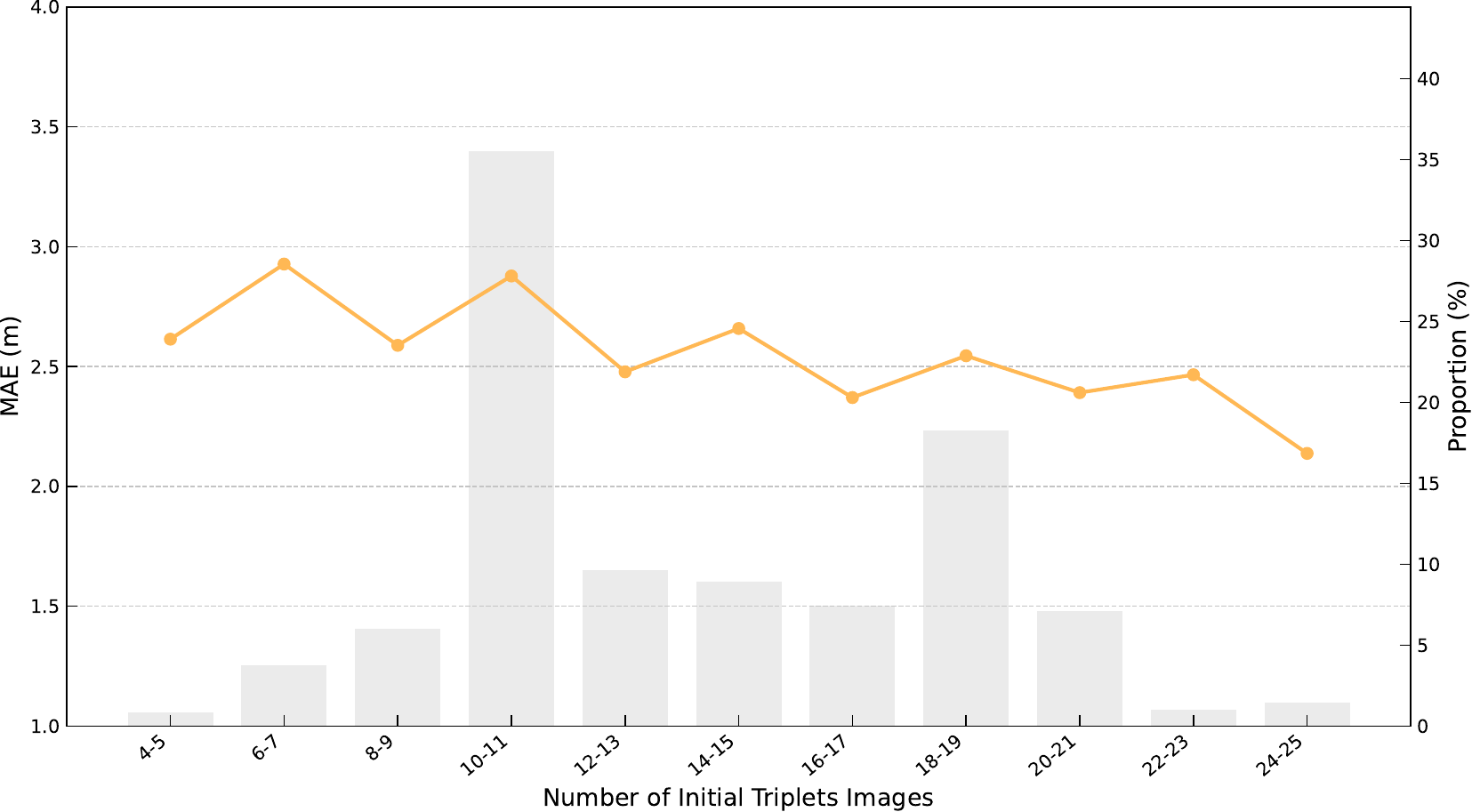}
    \caption{Evolution of the Mean Absolute Error (MAE) of SERA-H as a function of the number of initial image triplets (orange curve, left axis). The gray bars in the background represent the proportion of samples in the test dataset for each bin (right axis). The number of triplets refers to the number of unique acquisitions available before the duplication or truncation process used to reach the fixed input sequence length of 16 required by SERA-H}
    \label{fig:figure_8_MAE_by_nb_triplets}
\end{figure}

\begin{figure*}[!b]
    \centering
    \includegraphics[width=0.9\textwidth]{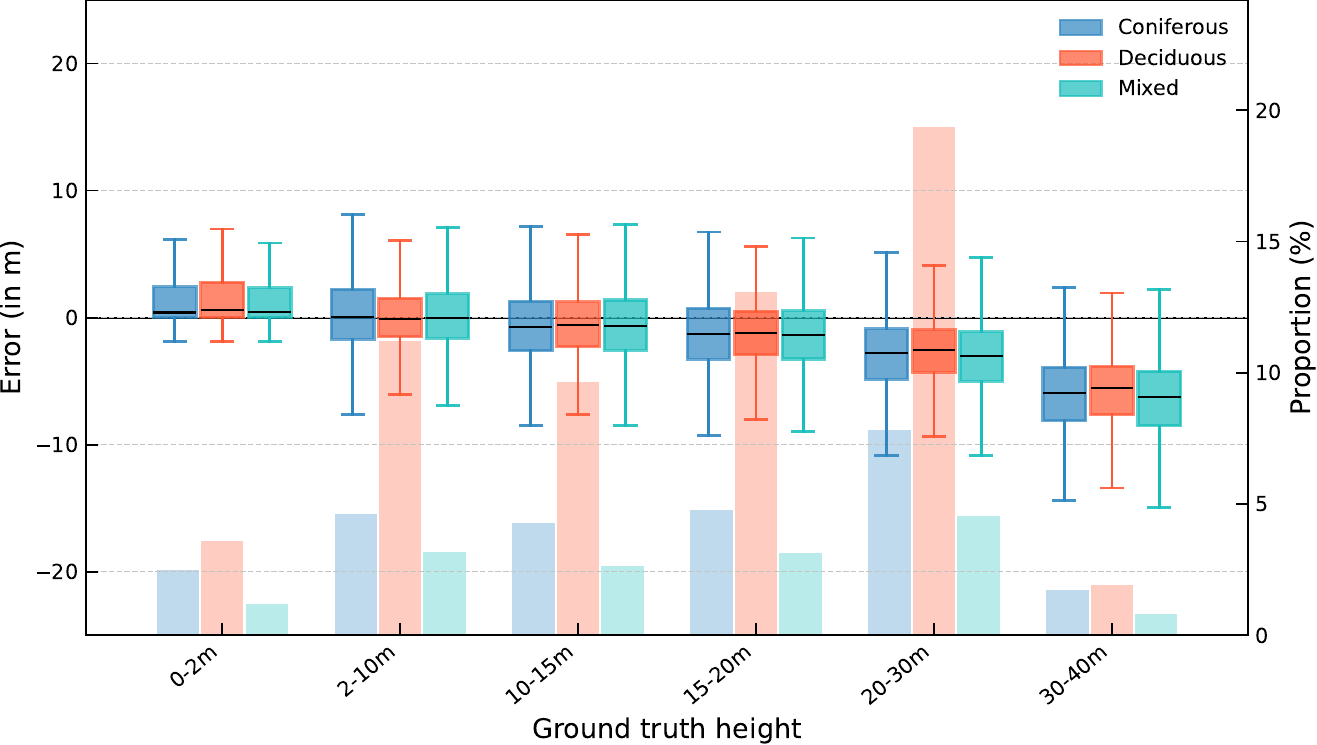}
    \caption{Distribution of the prediction error (prediction $-$ reference) per reference height class, separated by forest type (coniferous, deciduous, mixed) (left axis). The transparent bars in the background indicate the proportion of samples of each type in each class (right axis). The format follows that of \cref{fig:figure_5_boxplot}.}
    \label{fig:figure_9_boxplot_by_height_by_forest_type}
\end{figure*}

\section{Impact of Temporal Density on Model Performance}
\label{app:temporal_density}

To assess the robustness of SERA-H regarding data availability, we analyzed the Mean Absolute Error (MAE) as a function of the number of available image triplets  $(S2, S1_{\text{asc}}, S1_{\text{dsc}})$ before the duplication or truncation step (see \cref{sec:satellite_data})

As illustrated in \cref{fig:figure_8_MAE_by_nb_triplets}, the model performance improves slightly as the number of available triplets increases, which is expected since the model is provided with more temporal information. However, the MAE remains remarkably stable even for sequences with a low number of initial images. This stability demonstrates the model's capacity to effectively reconstruct canopy height even when temporal density is sparse.

\begin{figure*}[b]
    \centering
    \includegraphics[width=0.9\textwidth]{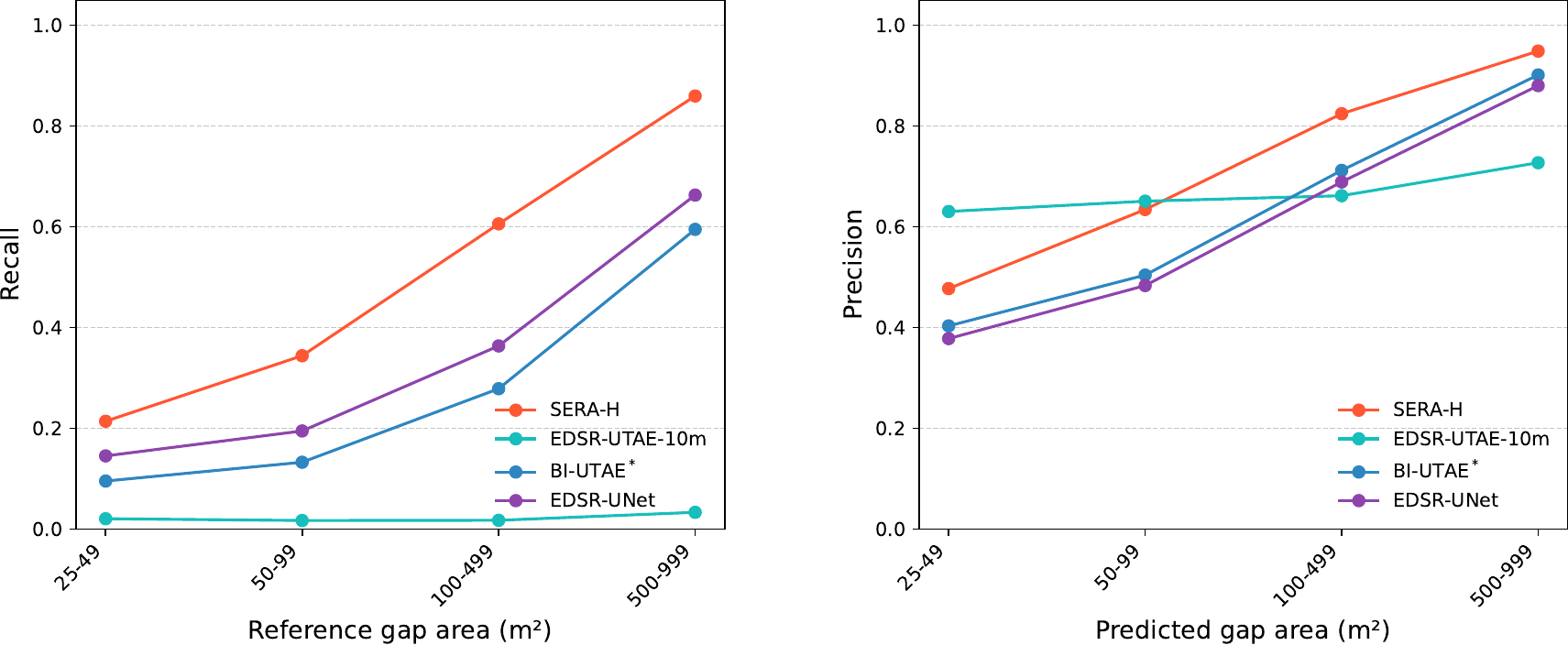}
    \caption{Canopy gap detection performance as a function of gap surface area. Recall (left) is reported per surface-area class of the reference gaps, and precision (right) per surface-area class of the predicted gaps. A predicted gap is counted as correctly detected when its IoU with a reference gap is at least 25~\%. SERA-H is compared against the BI-UTAE$^*$, EDSR-UNet, and EDSR-UTAE-10m variants.}
    \label{fig:figure_10_gap_detection}
\end{figure*}

\section{Saturation effect}
\label{app:saturation_by_forest_type}

We wanted to provide additional insight into the saturation effect. We therefore plotted the distribution of the error (prediction $-$ reference) per height class, separated by forest type (coniferous, deciduous and mixed, \cref{fig:figure_9_boxplot_by_height_by_forest_type}). Three trends emerge. First, the median error confirms a clear saturation effect: predictions are relatively unbiased up to about 20~m, then increasingly underestimate canopy height beyond, with the median bias reaching about $-5.9$~m in the 30--40~m class. Second, excluding the 0--2~m class, the interquartile range increases only very slightly with height: from 3.9 to 4.2~m for coniferous, 3.0 to 3.8~m for deciduous and 3.5 to 4.3~m for mixed stands, even as the median drops by nearly 6~m over the same range. Third, this behavior is largely shared across the three forest types. We note, however, that deciduous stands exhibit a slightly smaller dispersion.

The combination of a downward-shifting median with a nearly stable interquartile range indicates that this saturation manifests as a systematic and directional bias, the model underestimates tall canopies consistently rather than scattering its predictions. This saturation bias can be explained in two complementary ways. On the one hand, from a physical standpoint, optical and radar signals lose their sensitivity to height when the canopy becomes tall and dense. Optical reflectance (Sentinel-2) originates mainly from the upper canopy layer. Once the canopy is closed and the leaf area index is high, the signal no longer varies much with the actual height. Similarly, the short wavelength of C-band radar (Sentinel-1) penetrates only the top of the canopy, so that backscatter saturates rapidly with biomass. Beyond a certain height, the input signal therefore carries almost no information allowing the model to distinguish, for example, a 30~m stand from a 40~m one, which results in a systematic underestimation of tall trees. On the other hand, from a statistical standpoint, the 30--40~m class is markedly underrepresented compared to the others in the dataset. Trained under a loss function that minimizes the mean error, the model tends, when faced with an ambiguous signal and few examples, to predict toward the most frequent and therefore lower values. These two mechanisms compound each other: saturation makes the input signal ambiguous for tall canopies, and the scarcity of such examples leads the model to resolve this ambiguity toward lower values, producing a consistent bias rather than scattered predictions. The fact that this underestimation is shared across the three forest types supports a cause primarily related to height and to the sensor, rather than specific to a given stand type.

\section{Fine-scale reconstruction: gap and edge analysis}
\label{app:fine_scale_metrics}

While the qualitative and quantitative results (\cref{sec:results}) provide a good overview of the model's performance, they do not fully assess its ability to reconstruct fine-scale details at 2.5~m. To specifically test this point, we introduce two complementary metrics, based on canopy edge detection and canopy gap detection. For each, SERA-H is compared against the BI-UTAE$^*$, EDSR-UNet and EDSR-UTAE-10m variants, in order to isolate the respective contributions of the super-resolution module, the temporal series, and the high-resolution supervision.

\begin{figure*}[t]
    \centering
    \includegraphics[width=0.9\textwidth]{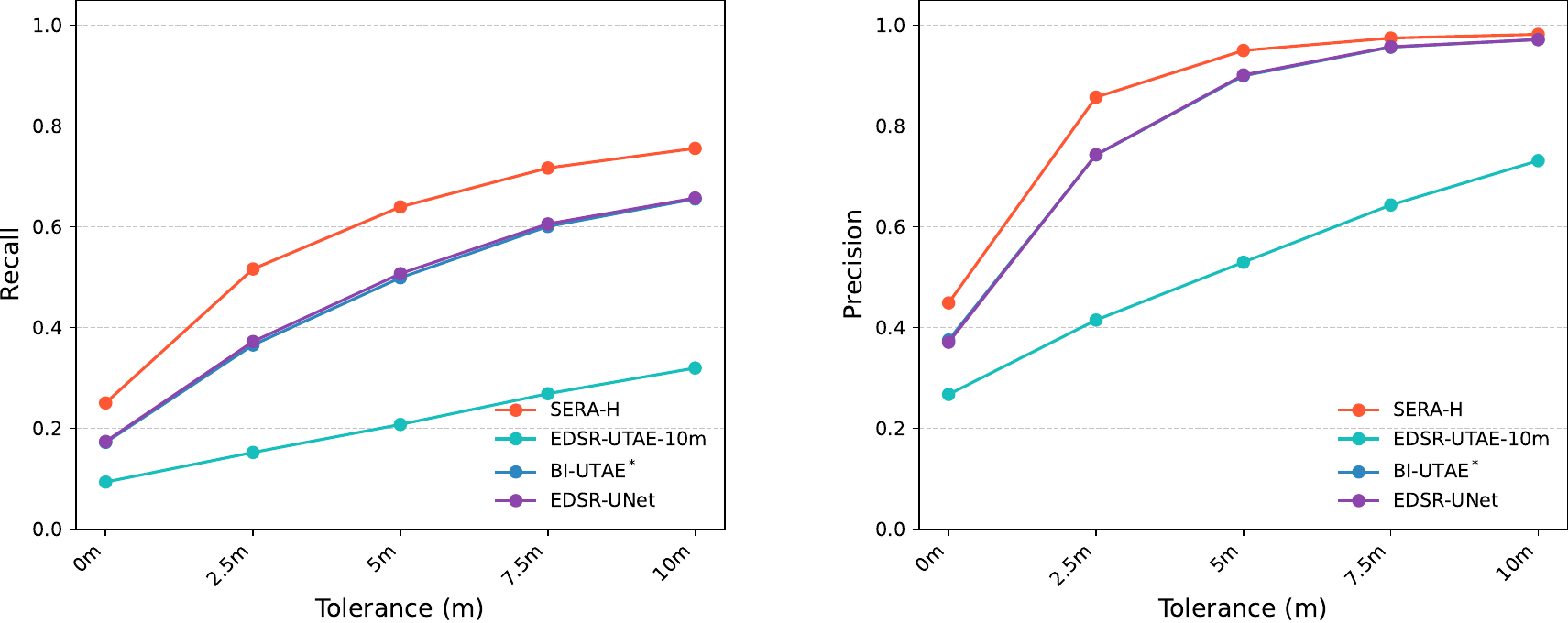}
    \caption{Canopy edge detection performance as a function of the spatial tolerance. Recall (left) and precision (right) are reported for increasing tolerance values (0 to 10~m), which allow an edge pixel to be counted as correct even when it is not exactly superimposed on the reference edge. SERA-H is compared against the BI-UTAE$^*$, EDSR-UNet, and EDSR-UTAE-10m variants.}
    \label{fig:figure_11_edge_detection}
\end{figure*}

\paragraph{Gap detection}
From the height maps, we produce a forest mask by applying a 2~m height threshold, as done for the Tree Cover IoU (\cref{sec:evaluation_metrics}). Within the resulting non-forest areas, we define as gaps the connected components with an area between 25 and 1000~m$^2$, i.e. the small canopy openings, excluding larger non-forest expanses that correspond to open land rather than gaps. Because a predicted gap may overlap several reference gaps, and vice versa, matching is performed at the level of connected groups: all predicted and reference gaps that overlap are grouped together, and a group is deemed correctly detected when the IoU between its reference and predicted regions is at least 25\%. On this basis, recall is the proportion of reference gaps correctly detected, and precision the proportion of predicted gaps that are correct, a gap counting as correct when its group passes the IoU criterion. We then report recall by surface-area class of the reference gaps, and precision by surface-area class of the predicted gaps in \cref{fig:figure_10_gap_detection}.

SERA-H shows a relatively high precision over a large part of the spectrum. It is 0.48 for the smallest gaps and increases steadily up to 0.95 for the largest ones. This indicates that the information provided by the model is reliable for medium- and large-sized gaps. For the smallest gaps, precision decreases as expected, but the result remains notable: although these gaps lie well below the native resolution of Sentinel, nearly half of those predicted at this scale are still correct. Recall, for its part, starts relatively low, at 0.21 for the 25--49~m$^2$ class and 0.34 for the 50--99~m$^2$ class. As these gaps are smaller than a Sentinel pixel, such information would not be expected to be captured at Sentinel's native resolution, recall nonetheless remains non-negligible, indicating that the model still recovers part of it. Recall then increases: above 100~m$^2$, it rises to 0.61 and reaches 0.87 for the 500--999~m$^2$ class, indicating that from this size onward the model misses only few gaps.

Regarding the comparison with the other models, SERA-H dominates all the models in the ablation. Only EDSR-UTAE-10m outperforms it in precision between 25 and 99~m$^2$. However, since its recall is so low in this range, this precision relies on only a very small number of gaps, which makes it not a truly significant result. The fact that SERA-H achieves both a higher recall and a higher precision than BI-UTAE$^*$, particularly for sizes below the resolution of Sentinel, highlights the contribution of the super-resolution module compared to bilinear interpolation, producing a map that is both more accurate and more complete. Likewise, the superior results of SERA-H over EDSR-UNet indicate that the additional information provided by the time series also improves the reconstruction, making it both more accurate and more complete, including at sizes below the resolution of Sentinel. Finally, the near-zero recall of EDSR-UTAE-10m tends to suggest that training the model on targets limited to 10~m degrades the reconstruction not only for gaps smaller than a Sentinel pixel, but also for larger gaps, confirming the value of training on high-resolution reference images.

\paragraph{Edge detection}
To extract forest edges, we again build a forest mask by thresholding the height maps at 2~m, as previously, and keep only the forest pixels located at the border of their zone. We then evaluate the degree of overlap between the predicted edges and the reference edges. In order not to penalize a slight spatial offset, we introduce a tolerance that allows an edge pixel to be counted as correct even when it is not exactly superimposed on the reference edge. Recall then measures the proportion of reference edge pixels detected within this tolerance, while precision measures the proportion of predicted edge pixels correctly matched to a reference edge pixel. By varying the tolerance, we assess in \cref{fig:figure_11_edge_detection} how finely the model localizes canopy edges.

As with the gaps, SERA-H also achieves a relatively high precision for edges across the entire tolerance range. Despite the strong constraint of a zero tolerance, which requires an exact superposition of the edge pixels, it already reaches 0.45. Precision then increases rapidly as soon as a small tolerance is allowed: 0.86 at 2.5~m, 0.95 at 5~m, and up to 0.98 at 10~m. This indicates that the edges predicted by the model are spatially very close to the reference, within a few meters, i.e. below the native resolution of Sentinel. Recall follows the same increasing trend. In the absence of tolerance, it starts relatively low but is not zero either (0.25): although this exact overlap of the edges remains very limited, it is nonetheless partly achieved by the model. Recall then increases markedly: it reaches 0.53 at 2.5~m, 0.65 at 5~m, and 0.77 at 10~m. Thus, as for the gaps, a substantial part of the edges is recovered at scales below the native resolution of Sentinel, confirming that the model localizes edges more finely than its input images would suggest.

For edges, the comparison with the other models confirms the hierarchy observed for gaps: SERA-H outperforms every ablation model at all tolerances, in both precision and recall. This again reflects the respective contributions of the super-resolution module and of the temporal series to high-resolution reconstruction, as well as the cost of training on targets limited to 10~m. As for the gaps, BI-UTAE$^*$ and EDSR-UNet yield very similar results, both clearly below SERA-H: super-resolution alone (EDSR-UNet) and the temporal series alone (BI-UTAE$^*$) reach a comparable level, and it is their combination that allows SERA-H to stand out. The way the two metrics evolve with tolerance further reveals two distinct advantages. In precision, SERA-H's margin is largest at fine tolerances ($+0.12$ at 2.5~m) and then narrows, the three best models converging around 0.96--0.98 at 10~m: all of them place the edges in the right area, but SERA-H localizes them more finely, within a few meters, below the resolution of Sentinel. In recall, by contrast, SERA-H retains its advantage at all tolerances (0.77 vs 0.66 for BI-UTAE$^*$ at 10~m), indicating that the other models not only localize some edges less precisely, but also miss part of them entirely. SERA-H thus produces maps that are both sharper (precision) and more complete (recall).

\paragraph{Summary}
In summary, the two metrics provide a quantitative assessment of SERA-H's fine-scale reconstruction at 2.5~m. The high precision values show that the information restored by the model, even below the resolution of Sentinel, is largely accurate. However, recall indicates that at too small a scale, part of the reference information is not recovered, although the recovered fraction remains non-negligible, which is consistent with the limitations discussed in the conclusion. These results, superior to those of the ablation variants, demonstrate the value, for fine-scale reconstruction, of combining the super-resolution module with the temporal series. The detail reconstructed by SERA-H thus appears real and spatially consistent with the reference, at a scale finer than that of its input images.

\section{Additional qualitative examples}
\label{app:additional_examples}

\begin{figure*}[!htbp]
    \centering
    \includegraphics[width=0.9\textwidth]{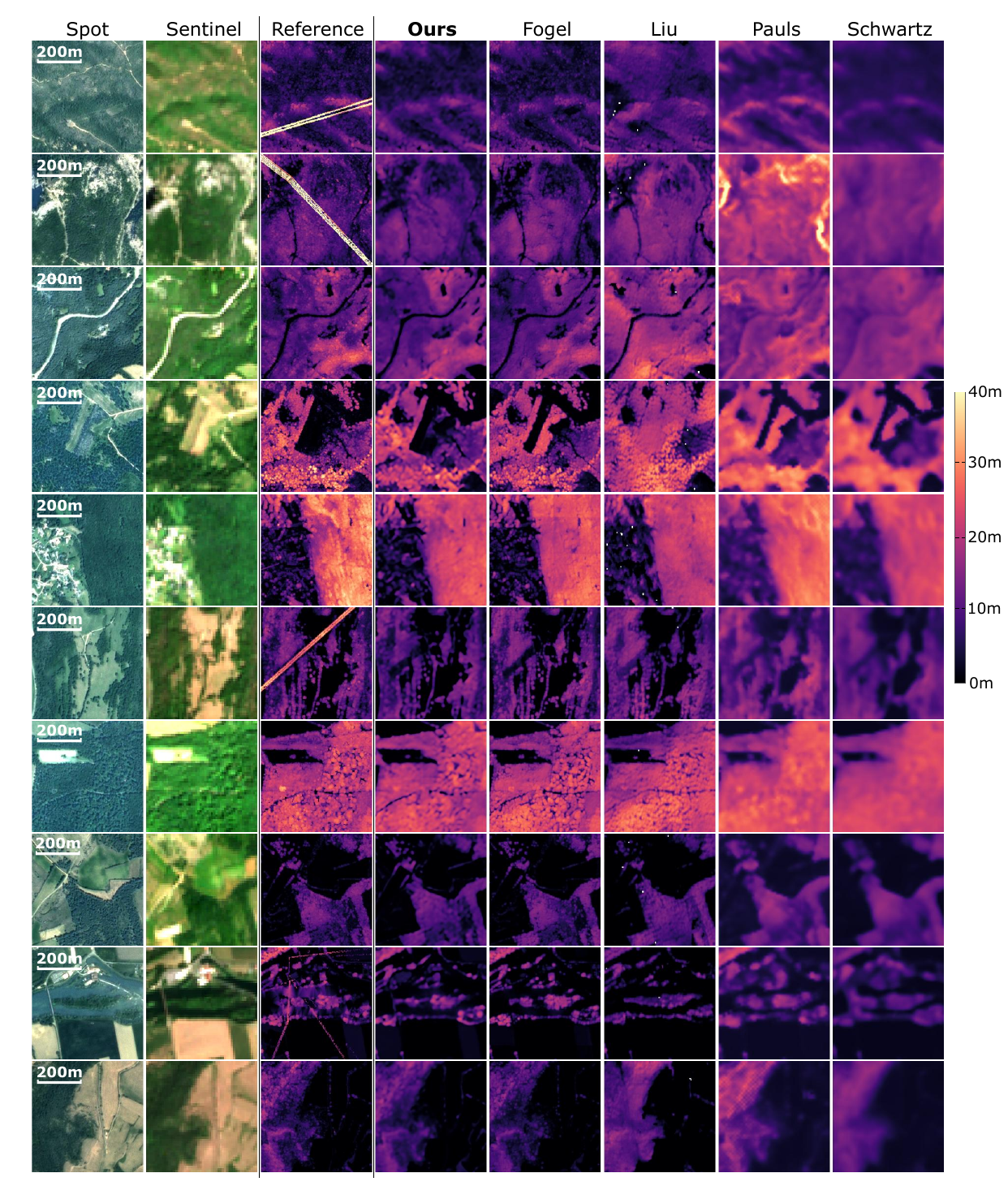}
    \caption{Ten additional examples randomly selected from the test set among tiles containing forest. Columns and height-color scale follow \cref{fig:figure_4_qualitative_results}}
    \label{fig:figure_12_qualitative_appendix}
\end{figure*}

To provide a broader and unbiased view of model behavior, we additionally report ten examples drawn at random from the test set among tiles containing forest (\cref{fig:figure_12_qualitative_appendix}). These randomly sampled sites confirm the trends observed in \cref{fig:figure_4_qualitative_results}: across contrasting forest types, SERA-H produces sharp canopy contours and recovers internal structural variations (tree-height gradients and forest gaps) that approach the individual-tree scale, clearly surpassing the other Sentinel-based products and remaining visually close to Fogel despite relying only on 10~m imagery. At the same time, this randomly sampled set also exposes the main failure modes of SERA-H: fine high-contrast linear features (e.g. elevated tracks, hedgerows) are smoothed, gap boundaries are softened, and the height of tall, dense stands is occasionally underestimated. These limitations are consistent with the native resolution of the Sentinel input and with the saturation effect discussed in \ref{app:saturation_by_forest_type}, and they affect all Sentinel-based models to a comparable or greater extent.

\clearpage
\bibliographystyle{elsarticle-harv}
\bibliography{references}

\end{document}